\documentclass[lettersize,journal]{IEEEtran}
\usepackage{amsmath,amsfonts,amssymb}
\usepackage{array}
\usepackage{booktabs}
\usepackage[caption=false,font=normalsize,labelfont=sf,textfont=sf]{subfig}
\usepackage{textcomp}
\usepackage{stfloats}
\usepackage{url}
\usepackage{graphicx}
\usepackage{cite}
\makeatletter
\def\@begintheorem#1#2{\@IEEEtmpitemindent\itemindent\relax\topsep 0pt\rmfamily\trivlist%
    \item[]\textit{\indent #1\ \textup{#2}:} \itemindent\@IEEEtmpitemindent\relax}
\def\@opargbegintheorem#1#2#3{\@IEEEtmpitemindent\itemindent\relax\topsep 0pt\rmfamily \trivlist%
    \item[]\textit{\indent #1\ \textup{#2}\ (#3):} \itemindent\@IEEEtmpitemindent\relax}
\makeatother
\usepackage{bm}
\usepackage{microtype}
\usepackage[table]{xcolor}
\usepackage{hyperref}
\hypersetup{hypertexnames=false}
\usepackage{xcolor}
\usepackage{tikz}
\usetikzlibrary{positioning,arrows.meta}
\usepackage{tabularx}
\usepackage{multirow}
\usepackage{bm}
\usepackage{bbding}
\hypersetup{hidelinks}
\usetikzlibrary{arrows.meta}
\makeatletter
\long\def\@makecaption#1#2{%
\ifx\@captype\@IEEEtablestring%
\footnotesize\bgroup\par\@IEEEtabletopskipstrut%
\setbox\@tempboxa\hbox{\normalfont\footnotesize {#1.}\nobreakspace\nobreakspace #2}%
\ifdim \wd\@tempboxa >\hsize%
\setbox\@tempboxa\hbox{\normalfont\footnotesize {#1.}\nobreakspace\nobreakspace}%
\parbox[t]{\hsize}{\normalfont\footnotesize\noindent\unhbox\@tempboxa#2}%
\else%
\hbox to\hsize{\normalfont\footnotesize\box\@tempboxa\hfil}%
\fi%
\par\addvspace{0.5\baselineskip}\egroup%
\@IEEEtablecaptionsepspace
\else%
\@IEEEfigurecaptionsepspace
\setbox\@tempboxa\hbox{\normalfont\footnotesize {#1.}\nobreakspace\nobreakspace #2}%
\ifdim \wd\@tempboxa >\hsize%
\setbox\@tempboxa\hbox{\normalfont\footnotesize {#1.}\nobreakspace\nobreakspace}%
\parbox[t]{\hsize}{\normalfont\footnotesize\noindent\unhbox\@tempboxa#2}%
\else%
\hbox to\hsize{\normalfont\footnotesize\box\@tempboxa\hfil}%
\fi\fi}
\makeatother

\begin{document}

\title{\textbf{\texttt{NOUS}}: Video-Driven 3D Human Reaction Generation via Observation-Reaction Mutual Steering}

\author{Yuan~Zhou, Luanyuan Dai, Yongzhi~Li, Shijie Hao, Xingyu~Zhu, Yi~Tan, Qingshan~Xu, Beier~Zhu, Richang~Hong,~\IEEEmembership{Senior Member,~IEEE}, and~Hanwang~Zhang,~\IEEEmembership{Senior Member,~IEEE}%
\thanks{\textcolor{magenta}{Code is available at \url{https://github.com/zhouyuan888888/NOUS}}}
\thanks{This work was supported by the National Natural Science Foundation of China (NSFC) under Grant Nos. 62671234 and 62676130.}
\thanks{Yuan Zhou, Luanyuan Dai, Yi Tan, and Hanwang Zhang are with the College of Computing and Data Science, Nanyang Technological University, Singapore 639798 (e-mail: yuan.zhou@ntu.edu.sg; luanyuan.dai@ntu.edu.sg; yi.tan@ntu.edu.sg; hanwangzhang@ntu.edu.sg).}%
\thanks{Xingyu Zhu is with the National University of Singapore, Singapore 119077 (e-mail: xyzhuxyz@mail.ustc.edu.cn).}%
\thanks{Qingshan Xu, Beier Zhu, and Yongzhi Li are with the School of Information Science and Technology, University of Science and Technology of China, Hefei 230026, China (e-mail: qingshan.xu@ustc.edu.cn; beier.zhu@ustc.edu.cn; wlee02320@mail.ustc.edu.cn).}%
\thanks{Richang Hong and Shijie Hao are with the School of Computer Science and Information Engineering (School of Artificial Intelligence), Hefei University of Technology, Hefei 230601, China (e-mail: hongrc.hfut@gmail.com; hfut.hsj@gmail.com).}%
\thanks{Corresponding authors: Qingshan Xu and Beier Zhu.}}

\markboth{IEEE Transactions on Multimedia,~Vol.~XX, No.~XX, 2026}%
{Zhou \MakeLowercase{\textit{et al.}}: \texttt{NOUS}: Reaction-Observation Mutual Steering for Video-Driven 3D Human Reaction Generation}


\maketitle
\bstctlcite{BSTcontrol}

\begin{abstract}
Video-driven 3D human reaction generation aims to synthesize 3D human motion in response to the action observed in a video, playing an important role in interactive multimedia systems and embodied agents. 
Yet reaction motions generated by current methods often fail to match what the observed video calls for. We observe that one factor behind this failure is relational distortion in the correspondence between visual observations and reactions: videos lying close in the visual space may correspond to entirely different motions in the reaction space, which misleads the model into generating reactions inconsistent with the conditioning video. This motivates us to propose a new observatioN-reactiOn mUtual Steering (\texttt{NOUS}) framework that enables mutual steering between the video and motion modalities. It first performs Motion Feedback Steering (MFS), equipping the frozen pretrained video encoder with a lightweight rectification modulator and training the modulator with a relational margin loss that pulls each video embedding toward the motion prototype of its own category and away from those of other categories. In this way, the misaligned correspondence between visual observations and reactions can be calibrated. \texttt{NOUS} then applies Observation-Guided Refinement (OGR), which in turn exploits the rectified observations to further refine the generated reactions and enhance their quality.
The results on the ViMo dataset demonstrate that \texttt{NOUS} improves the quality of reaction motion while incurring negligible computational overhead at inference. Also, \texttt{NOUS} yields consistent gains across four pretrained video encoders, showing its good compatibility.
\end{abstract}


\section{Introduction}
\label{sec:intro}

3D human reaction generation aims to synthesize 3D human motion in response to environmental stimuli, such as actions of surrounding agents, a capability that matters for human-robot interaction, virtual reality, and embodied AI. Recent progress~\cite{xu2024regennet,gohar2024intermask,tan2025think,wang2025timotion} has largely concerned human-human interaction, where the stimulus is provided as an explicit 3D representation of another person's motion; however, such 3D human motion information is rarely available in real-world settings, whereas video observations of human interactions are easier to obtain~\cite{tang2025video}. Deploying these methods in practice thus entails a two-stage pipeline that needs to recover 3D motion from raw footage and then synthesizes the reaction on top of it, which is cumbersome and inefficient. To overcome this issue, Yu et al.~\cite{hero} recently introduced Video-driven 3D Human Reaction Generation (VHRG), which targets generating plausible 3D human reactions directly from video sequences.

Despite its substantial potential in real-world applications, VHRG remains challenging. We find that the current method often reacts incorrectly to the observed video. As shown in Fig.~\ref{fig:rd}~(a), when the video shows a person opening their arms for a hug, the model answers with a wave. The video is observed, yet the model produces a reaction other than the one
it calls for, suggesting that something breaks between observing and reacting. This motivates us to ask two natural questions:
\begin{itemize}
\item \textbf{RQ1: What limits the effectiveness of the conditioning video in guiding 3D human reaction generation?}
\item \textbf{RQ2: How can such a limitation be addressed so that generated reactions better match their conditioning~video?}
\end{itemize}

\noindent 
Our analysis of \textbf{RQ1} reveals a distortion in the correspondence between visual observations and reaction motions. Videos that lie close in the visual embedding space may correspond to entirely different motions in the reaction space, a phenomenon we term \emph{relational distortion}. One reason is that pretrained video encoders are usually not explicitly optimized to capture video-reaction correspondence. With paired video-motion data remaining scarce, existing methods often freeze such encoders, leaving the distortion uncorrected in the conditioning signal. Conditioned on such a distorted signal, the generator is prone to producing reaction motion inconsistent with the interaction intent conveyed by the conditioning video.

\begin{figure}[t!]
    \centering  \includegraphics[width=1\linewidth]{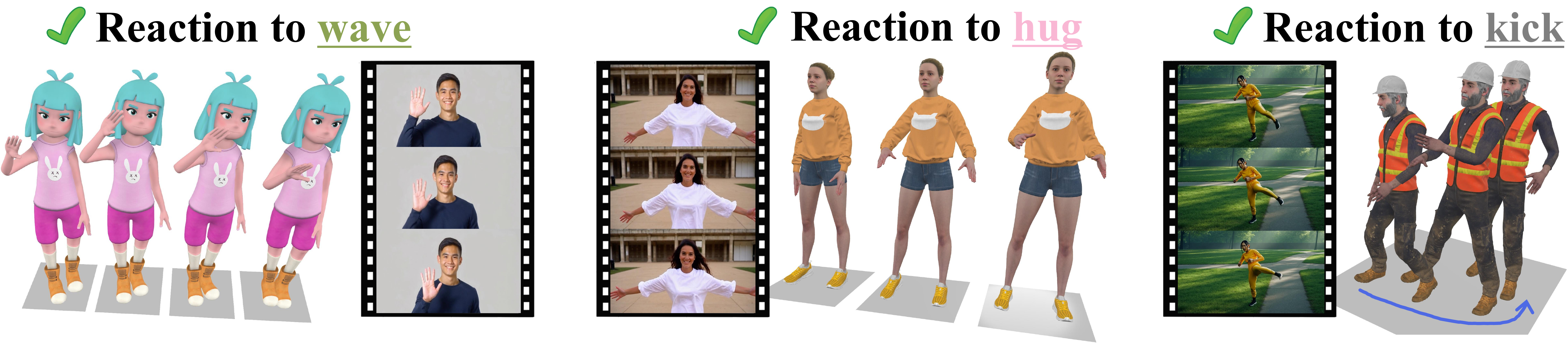}
    \caption{\textbf{Video-driven 3D human reaction generation with \texttt{NOUS}.} Given a video of a behavior such as waving, hugging, or kicking, our \texttt{NOUS} synthesizes a plausible 3D human motion in response to the observed action.}
    \label{fig:teaser}
    \vspace{-0.2cm}
\end{figure}

To address \textbf{RQ2}, we propose observatioN-reactiOn mUtual Steering (\texttt{NOUS}), a new framework for VHRG that overcomes relational distortion through two design principles. First, the video encoder stays frozen, since paired video-motion data are far scarcer than large-scale video data used in pretraining, and keeping it frozen preserves its pretrained visual knowledge. The correction therefore acts on embeddings the encoder outputs. Second, we draw the target of this correction from the reaction side, because targets computed in the video space would inherit the distortion we aim to repair, whereas paired reaction motions directly indicate which videos call for the same reaction.

Following these two design principles, \texttt{NOUS} operates in two stages. The first, Motion Feedback Steering (MFS), computes a non-parametric prototype for each reaction category from the motion space, then trains a lightweight delta-rectification modulator on top of the frozen encoder, correcting its output embeddings by pulling each toward its own prototype and away from the others under a relational margin loss. The second, Observation-Guided Refinement (OGR), pairs the rectified embedding with the original to form a joint condition and uses it to further refine the model's own predictions. On the ViMo dataset, \texttt{NOUS} reduces FID from 0.427 to 0.328 (a 23\% improvement) while adding only 6.7\% more parameters (31.2\,M to 33.3\,M) and 2.8\% more FLOPs (36.1\,G to 37.1\,G) per sample at inference. Our \texttt{NOUS} also presents good compatibility, achieving consistent gains across four video encoders, TC-CLIP~\cite{kim2024leveraging}, VideoMAE~\cite{wang2023videomae}, ViCLIP~\cite{wang2023internvid}, and X-CLIP~\cite{ni2022expanding}.

Our contributions are threefold:
\begin{enumerate}
    \item We identify relational distortion in video-driven reaction generation, whereby videos lying close in the visual space may correspond to entirely different reaction motions, and show that reducing this distortion consistently improves reaction synthesis. This finding reveals a representation-level mismatch in video conditioning and motivates extracting visual observations that better preserve the video-reaction correspondence.
    
    \item We propose \texttt{NOUS}, where MFS rectifies the distorted correspondence by pulling each extracted visual observation toward its own motion prototype and away from the others via a lightweight rectification modulator, and OGR in turn exploits the rectified observations to refine the generated reactions, thus improving their quality.
    
    \item Extensive experiments on the ViMo dataset show that our \texttt{NOUS} consistently improves reaction quality across different pretrained video encoders while incurring negligible computational cost at inference.
    
\end{enumerate}

\section{Related Work}
\label{sec:rela}

\subsection{3D Human Motion Generation} 
3D human motion generation aims to create realistic and expressive 3D human motion conditioned on user-provided instruction. Prior studies employ a wide spectrum of generative paradigms to model human motion sequences. For example, \cite{wang2020learning,xu2023actformer,yan2019convolutional} were built upon GANs \cite{goodfellow2014generative}, \cite{dai2024motionlcm,huang2024stablemofusion,kim2023flame,kong2023priority,shafir2023human,tseng2023edge,yuan2023physdiff,zhou2024emdm} utilized diffusion models \cite{ho2020denoising,Zhu_2025_ICCV,sohl2015deep} to achieve high-quality synthesis, and \cite{athanasiou2022teach,cervantes2022implicit,guo2020action2motion,guo2022generating,petrovich2021action,petrovich2022temos} resorted to VAEs \cite{kingma2013auto} to model stochastic relationships between conditions and motions. In addition, \cite{gong2023tm2d,guo2022tm2t,jiang2023motiongpt,t2m-gpt,zhang2024motiongpt,zhong2023attt2m} formulated human motion generation in an autoregressive manner, whereas masked generative modeling \cite{chang2022maskgit} was employed in recent methods \cite{momask,gohar2024intermask,pinyoanuntapong2024controlmm,pinyoanuntapong2024mmm} to enable high-fidelity human motion synthesis with improved efficiency. Beyond textual descriptions, the conditioning signal spans a broad range of modalities, such as audio-driven motion synthesis~\cite{qi2024emotiongesture,wang2024crossmodal,sun2020deepdance}, and scene-aware generation~\cite{gao2025jointly}. While our \texttt{NOUS} also leverages masked generative modeling, it tackles a fundamentally different task--video-driven human reaction generation~\cite{hero}--which requires the model to infer interaction intent from video sequences and to generate plausible reaction motion that directly responds to the observed visual content.

\subsection{3D Human Reaction Generation} 
3D human reaction generation requires models to infer interaction intent from an input video and synthesize a corresponding reaction, distinguishing it from human motion synthesis that typically relies on explicit conditioning signals. Existing approaches mainly focus on human-human interaction, where the goal is to predict the reactor’s motion solely conditioned on the actor’s 3D pose\cite{chopin2023interaction,ghosh2024remos,liu2023interactive,xu2024regennet}. Diffusion models have been adopted in \cite{xu2024regennet,ghosh2024remos} to model human-human interaction, \cite{chopin2023interaction} proposed fully capturing spatio-temporal dependencies to enhance interaction coherence, and \cite{liu2024physreaction} introduced a forward-dynamics-guided imitation strategy for generating physically plausible human reactions. 
These methods model temporal motion dynamics but differ from 3D human motion prediction~\cite{shi2023multi,yu2024divdiff} by conditioning the reactor's motion on the partner's behavior rather than solely on its own motion history.
Since videos serve as an important visual modality for machine perception, Yu et al. \cite{hero} recently proposed a video-driven 3D human reaction generation task, aiming to enable models to synthesize natural, plausible reaction motions that directly interact with video sequences. Our work builds upon~\cite{hero} and goes a step further by moving beyond~its one-way pipeline to recast the framework into observation-reaction mutual steering, which further improves generation quality.

\section{Relational Distortion Analysis}
\label{sec:rda}
This section addresses \textbf{RQ1} in three steps: defining relational distortion, quantifying it empirically, and discussing why it may arise in the current VHRG pipeline.

\textbf{Relational Distortion.}
\label{def:rd}
Let $\bm{\mathcal{E}}=\{\bm{e}_i\}_{i=1}^{N}$ be the observation embeddings a frozen video encoder assigns to $N$ videos, and $\{y_i\}_{i=1}^{N}$ the  categories of reaction motions those videos call for. We say that relational distortion occurs in $\bm{\mathcal{E}}$ if relations among observation embeddings are inconsistent with their reaction types, i.e., the similarity score $s(\bm{e}_i,\bm{e}_j)$ is nearly as high when $y_i \neq y_j$ as when $y_i = y_j$.

Relational distortion is essentially a rift between two spaces: the visual space where the frozen encoder embeds what it sees, and the motion space where reactions are represented. The generator takes its condition from one and generates in the other, so once the two are misaligned, the correspondence between \emph{what is observed} and \emph{how one should react} is obscured, hindering the generation of high-quality reaction motion.

\begin{figure}[t!]
    \centering  \includegraphics[width=0.95\linewidth]{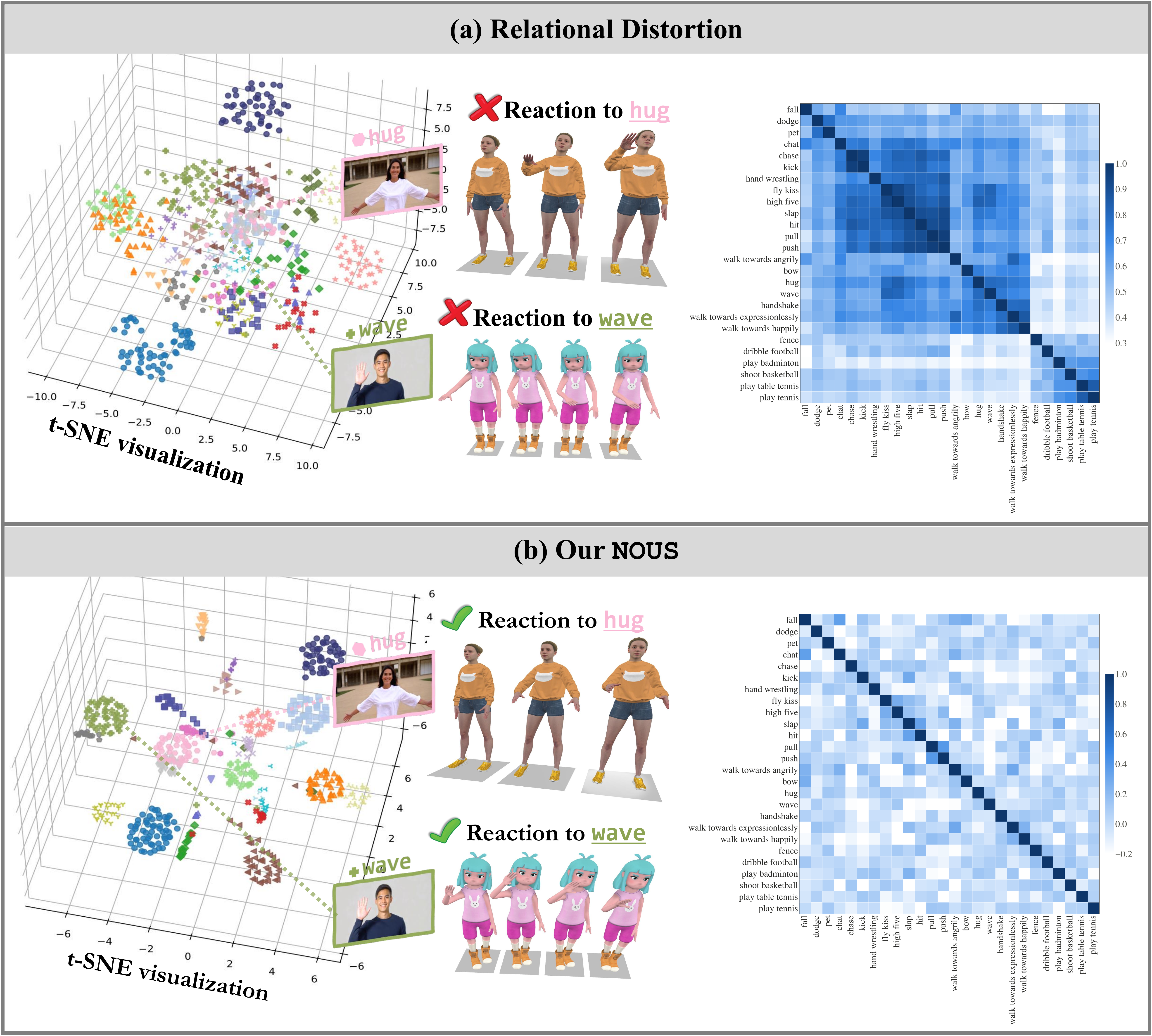}
    \vspace{-0.2cm}
    \caption{\textbf{t-SNE visualization of extracted visual observations with respect to reaction categories.} Each panel shows the extracted visual embeddings colored by reaction category, the reactions generated for the ``hug'' and ``wave'' videos, and the cosine similarities between the mean visual embeddings of reaction categories. (a)~The baseline shows overlapping categories, high off-diagonal similarities, and reactions inconsistent with the video content. (b)~\texttt{NOUS} yields separated clusters, lower off-diagonal similarities, and correct reactions.}
    \vspace{-0.3cm}
    \label{fig:rd}
\end{figure}

\begin{figure}[t]
\centering
\begin{tikzpicture}[>=latex,semithick,lat/.style={circle,draw=gray!70,dashed,fill=gray!10,minimum size=8mm,inner sep=1pt},obs/.style={circle,draw,fill=gray!12,minimum size=8mm,inner sep=1pt},new/.style={circle,draw=blue,text=blue,fill=blue!8,minimum size=8mm,inner sep=1pt},hyp/.style={->,gray!70},flow/.style={->,black},method/.style={->,blue},supervision/.style={->,blue}]
\node[obs] (v) at (0,0.75) {$\bm{V}$};
\node[lat] (zv) at (1.3,1.5) {$\bm{Z}_{\text{vis}}$};
\node[lat] (zi) at (1.3,0) {$\bm{Z}_{\text{int}}$};
\node[obs] (e) at (3.4,1.5) {$\bm{e}$};
\node[obs] (m) at (5.4,0) {$\bm{m}$};
\node[new] (eh) at (5.4,1.5) {$\bm{e}'$};
\draw[hyp] (zv) -- (v);
\draw[hyp] (zi) -- (v);
\draw[flow] (zv) -- node[above,font=\scriptsize]{Encoder} node[below,font=\scriptsize]{(frozen)} (e);
\draw[hyp] (zi) -- node[below,font=\scriptsize]{Reaction} (m);
\draw[method] (e) -- node[above,font=\scriptsize]{Modulator} (eh);
\draw[supervision] (zi) -- node[below,sloped,pos=0.62,align=center,font=\scriptsize]{Rectification\\[-1pt](training only)} (eh);
\end{tikzpicture}
\vspace{-0.2cm}
\caption{\textbf{Conceptual illustration of how relational distortion may arise and how \texttt{NOUS} mitigates it.} Arrows denote dependencies among variables, dashed circles denote latent factors, and what \texttt{NOUS} induces is drawn in blue. The video $\bm{V}$ carries generic visual content $\bm{Z}_{\text{vis}}$ and interaction-relevant cues $\bm{Z}_{\text{int}}$. The reaction motion $\bm{m}$ depends mainly on $\bm{Z}_{\text{int}}$, yet a frozen encoder pretrained without reaction-side supervision may emphasize $\bm{Z}_{\text{vis}}$ in its embedding $\bm{e}$. \texttt{NOUS} instead uses reaction-side supervision to train a modulator that maps $\bm{e}$ to $\bm{e}'$, connecting it to $\bm{Z}_{\text{int}}$ as well as $\bm{Z}_{\text{vis}}$.}
\vspace{-0.3cm}
\label{fig:scm}
\end{figure}

\vspace{-0.2cm}
\subsection{Empirical Analysis}
\label{subsec:diag}
In Fig.~\ref{fig:rd}~(a), we examine visual observations the baseline~\cite{hero} conditions on, where each point is a visual embedding colored by the reaction category its video calls for. The t-SNE visualization \cite{van2008visualizing} shows that points from different classes are interleaved rather than separated, indicating that videos lying close in the visual space often correspond to entirely different motions in the reaction space. This mismatch between the visual space and the reaction space provides a plausible mechanism by which the generator may be misled into producing the wrong reaction. The two examples further illustrate that the baseline~\cite{hero} generates reactions inconsistent with the content of both the ``hug'' and ``wave'' videos. The similarity matrix also quantifies this pattern at the category level by computing pairwise cosine similarities between the mean visual embeddings of reaction categories, where high off-diagonal entries suggest spurious correlations among reaction categories in the visual space. The supplementary analysis also shows a small gap between intra- and inter-class $\ell_2$ distances before rectification, suggesting that reaction categories are poorly separated in the visual space. As shown in Fig.~\ref{fig:rd}~(b), our method rectifies visual embeddings to form more distinct clusters and reduce off-diagonal similarities, thereby reducing interference from spurious correlations and helping the generator produce reaction motion that better matches the content of the conditioning video.

\subsection{Why Relational Distortion Arises}
\label{subsec:causal}

Fig.~\ref{fig:scm} illustrates how relational distortion may arise, where arrows denote dependencies among variables and dashed circles denote latent factors. We view the information carried by a video $\bm{V}$ as comprising two latent factors: generic visual content $\bm{Z}_{\text{vis}}$ and interaction-relevant cues $\bm{Z}_{\text{int}}$. A plausible reaction motion $\bm{m}$ depends primarily on $\bm{Z}_{\text{int}}$; in contrast, a frozen video encoder pretrained without reaction-side supervision is optimized to capture generic visual cues and may therefore emphasize $\bm{Z}_{\text{vis}}$ over $\bm{Z}_{\text{int}}$ in its embedding $\bm{e}$. Consequently, the visual observations extracted from a video may not provide sufficient cues to determine the appropriate reaction, hindering the generator from producing high-quality reaction motion.

This mismatch can occur systematically because conventional video pretraining objectives rarely encourage models to capture the relationship between a video and the reaction it calls for. One direct solution is to pretrain or finetune the video encoder using paired video-reaction data. However, such data are far scarcer than text-motion or text-video pairs commonly used in related generation tasks. Fully finetuning a pretrained video encoder in this limited-data regime may compromise its transferable knowledge acquired during pretraining.

\noindent\textbf{Remark.}
The preceding analysis highlights the need to rectify extracted visual observations to improve their correspondence with reaction motions. We realize this objective through motion feedback supervision, discouraging high similarity between observation embeddings paired with motions from different categories. Fig.~\ref{fig:scm} conceptually illustrates this rectification as a bridge toward $\bm{Z}_{\text{int}}$, injecting reaction-relevant information into the conditioning embedding. The blue path in Fig.~\ref{fig:scm} illustrates our design: the video encoder is frozen, while a lightweight modulator learns to map $\bm{e}$ to $\bm{e}'$ under motion-side supervision, thereby linking the rectified visual embeddings to both $\bm{Z}_{\text{int}}$ and $\bm{Z}_{\text{vis}}$. Optimizing only the lightweight modulator reduces trainable parameters and helps mitigate the impact of limited data, as video-reaction pairs are costly to collect.

\section{Methodology}
\label{sec:metho}

\subsection{Preliminaries}
\label{subsec:prel}

\textbf{VHRG.} 
Video-driven Human Reaction Generation (VHRG) aims to synthesize a human reaction that responds to the action observed in a video sequence. The training set consists of $N$ pairs $\{(\bm{v}_i,\bm{r}_i)\}_{i=1}^{N}$, where $\bm{v}_i \in \mathbb{R}^{F \times H \times W \times 3}$ is a video of $F$ frames at resolution $H \times W$, and $\bm{r}_i \in \mathbb{R}^{P_i \times D}$ is a reaction comprising $P_i$ poses of dimension~$D$. Each video corresponds to a ground-truth reaction type $y_i \in \{1,\dots,K\}$. Following~\cite{hero}, we adopt the 263-dimensional pose representation of~\cite{guo2022generating}.
\smallskip

\textbf{RVQ-VAE.} 
Reaction motions are tokenized by an RVQ-VAE. A motion encoder $\mathsf{Enc}_{\text{m}}$ maps $\bm{r}_i\in \mathbb{R}^{P_i \times D}$ to a latent sequence of length $M_i$, and Residual Vector Quantization~\cite{zeghidour2021soundstream} discretizes it with $L+1$ cascaded quantizers ($L$\,=\,5 here), each encoding the residual left by the previous one. This yields $L+1$ token sequences $\{\bm{s}_i^l\}_{l=1}^{L+1}$, each of length $M_i$: the first layer $\bm{s}_i^1$ carries the coarse motion, and each higher layer adds progressively finer details. A motion decoder $\mathsf{Dec}_{\text{m}}$ is also employed to map the summed quantized latents back to a reaction. We train the model with the standard reconstruction and commitment loss used in~\cite{momask,hero}.
\smallskip

\textbf{Masked Motion Modeling.} 
We model the first-layer motion tokens $\bm{s}_i^1$ with a masked prediction objective~\cite{chang2022maskgit,momask}, following Yu et al.~\cite{hero}. During training, we mask a random subset of positions to obtain a corrupted sequence $\hat{\bm{s}}_i^1$, and a Transformer $\mathsf{T}$ learns to recover the clean sequence conditioned on the visual embedding $\bm{e}_i$ extracted from a video $\bm{v}_i$:
\begin{equation}
L_{\text{mask}}
=-\mathbb{E}_{i}\log \mathbb{P}_{\mathsf{T}}\Bigl(\bm{s}_i^1 \Bigm| \hat{\bm{s}}_i^1,\, \bm{e}_i\Bigr).
\label{eq:mask}
\end{equation}
We also adopt the cosine masking schedule, remasking strategy, and Transformer architecture used in~\cite{hero}. At inference, the reaction generation process starts from an all-masked sequence and iteratively fills in predicted motion tokens.

\begin{figure*}[t!]
    \centering    
    \includegraphics[width=0.9\linewidth]{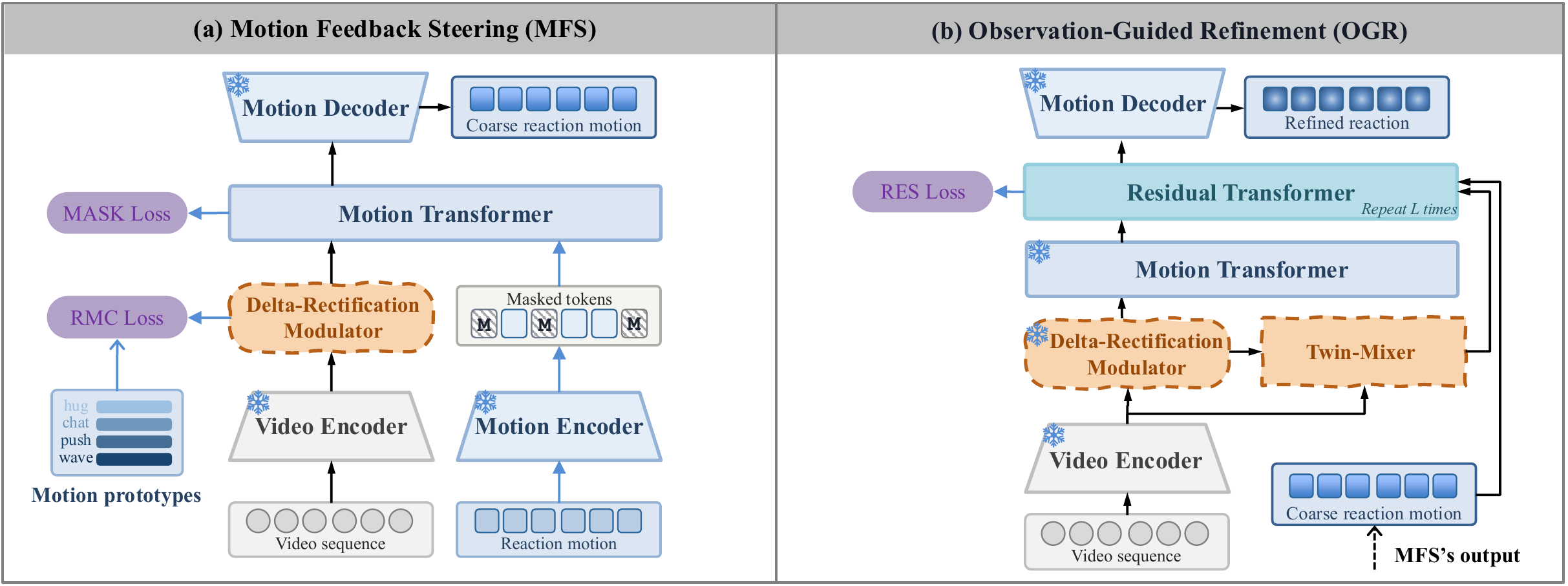}  
    \vspace{-0.2cm}
    \caption{\textbf{Schematic of \texttt{NOUS}.} (a) Stage-1 (MFS): the delta-rectification modulator learns to rectify embeddings from the frozen video encoder under the relational margin constraint (RMC) loss, while the motion Transformer is trained on the rectified embeddings with the masked prediction (MASK) loss. (b) Stage-2 (OGR): the residual Transformer and the Twin-Mixer are trained with the residual refinement (RES) loss to refine the coarse reaction motion from Stage-1. Snowflakes \Snowflake\ mark frozen modules, and ``\texttt{M}'' denotes a masked motion token.}
    \label{fig:NOUS}
\end{figure*}

\subsection{\texttt{NOUS}: Observation-Reaction Mutual Steering}
\label{subsec:MuSteerNet}

We propose \texttt{NOUS} to mitigate the relational distortion issue discussed in Sec.~\ref{sec:rda}. As shown in Fig.~\ref{fig:NOUS}, \texttt{NOUS} is a two-stage framework that enables mutual steering between visual observations and reaction motions. In Stage-1, Motion Feedback Steering (MFS) rectifies visual observations leveraging feedback from reaction motions. In Stage-2, Observation-Guided Refinement (OGR) combines the original and rectified observations to refine the coarse motions generated in Stage-1. We introduce these two stages in turn below.

\smallskip
\noindent\emph{1) Stage-1: Motion Feedback Steering}
\smallskip

Following Yu et al.~\cite{hero}, we train a base Transformer $\mathsf{T}_{\text{b}}$ using the objective in Eq.~\eqref{eq:mask}. We then build one prototype per reaction category by averaging token embeddings from $\mathsf{T}_{\text{b}}^{\text{emb}}$:
\begin{equation}
\bm{p}_{k}=\frac{1}{|\bm{\mathcal{I}}_k|}\sum_{i\in\bm{\mathcal{I}}_k}\frac{1}{M_i}\sum_{n=1}^{M_i}\mathsf{T}_{\text{b}}^{\text{emb}}\bigl(\bm{s}_{i,n}^{1}\bigr),
\label{eq:proto}
\end{equation}
where $\bm{s}_{i,n}^{1}$ is the $n$-th first-layer token of reaction $\bm{r}_i$, and $\bm{\mathcal{I}}_k=\{i \mid y_i=k\}$ indexes the samples in reaction category $k$. The resulting prototypes are normalized as $\bm{p}_{k}\leftarrow\bm{p}_{k}/\lVert\bm{p}_{k}\rVert_2$.

With these prototypes, we jointly train a motion Transformer $\mathsf{T}_{\text{m}}$ and a delta-rectification modulator $\mathsf{R}$. The modulator maps the embedding $\bm{e}_i=\mathsf{Enc}_{\text{v}}(\bm{v}_i)$ from the frozen video encoder to $\bm{e}_i'=\mathsf{R}(\bm{e}_i)$, which replaces $\bm{e}_i$ as the conditioning input to $\mathsf{T}_{\text{m}}$. We optimize $\mathsf{R}$ with both $L_{\text{mask}}$ and a relational margin constraint $L_{\text{rmc}}$, which pulls $\bm{e}_i'$ toward the prototype of reaction category $y_i$ and pushes it away from the other prototypes:
\begin{equation}
L_{\text{rmc}}
=-\mathbb{E}_{i}\log
\frac{e^{\mu\bigl[\cos(\bm{e}_i',\,\bm{p}_{y_i})-\varepsilon\bigr]}}
{e^{\mu\bigl[\cos(\bm{e}_i',\,\bm{p}_{y_i})-\varepsilon\bigr]}+\sum_{j\neq y_i}e^{\mu\cos(\bm{e}_i',\,\bm{p}_j)}},
\label{eq:rmc}
\end{equation}
where $\cos(\cdot,\cdot)$ denotes cosine similarity, and $\mu$ and $\varepsilon$ are the scaling factor and additive cosine margin, respectively. With $\bm{p}_{y_i}$ as the positive and all other prototypes as negatives, $L_{\text{rmc}}$ enlarges the margin between extracted visual observations belonging to different reaction categories.

\begin{figure}[t]
\centering
\includegraphics[width=0.68\linewidth]{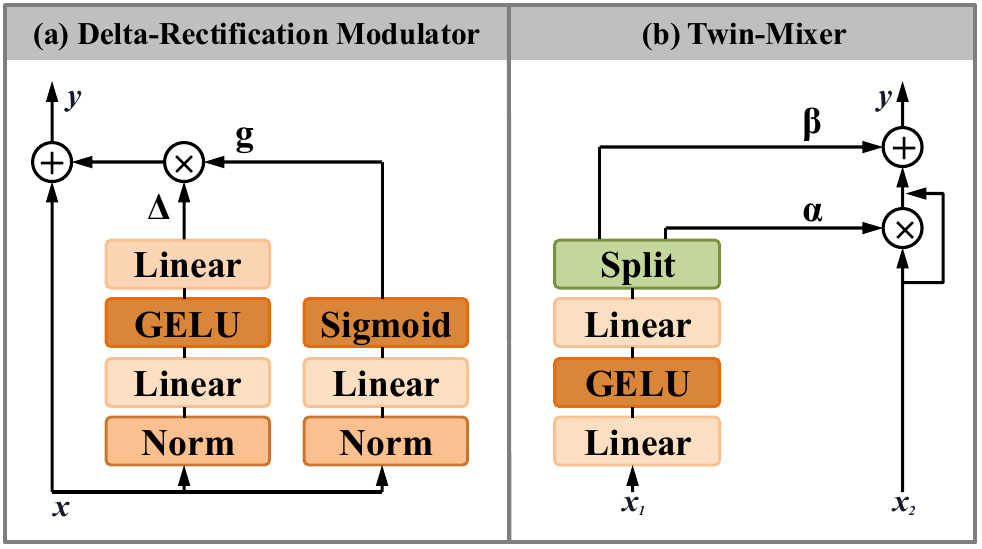}
\vspace{-0.2cm}
\caption{\textbf{Architectures of the delta-rectification modulator (a) and Twin-Mixer (b).} ``Norm'' and ``GELU'' denote LayerNorm and the Gaussian Error Linear Unit. In (a), $\bm{x}$ and $\bm{y}$ are the modulator input and output, while $\bm{\Delta}$ and $\bm{g}$ denote the residual correction term and the gate term. In (b), the inputs are $(\bm{x}_1,\bm{x}_2)=(\bm{e}_i',\bm{e}_i)$, and the output $\bm{c}_i=\bm{y}$ is the twin visual conditioning embedding used in Eq.~\eqref{eq:res}; $\bm{\beta}$ and $\bm{\alpha}$ denote the shift and scale adjustment.}
  \label{fig:arch}
\end{figure}

\textbf{Delta-Rectification Modulator.} Fig.~\ref{fig:arch}~(a) presents the architecture of our delta-rectification modulator $\mathsf{R}$. It adopts a gated residual design $\mathsf{R}(\bm{x})=\bm{x}+\bm{g}\odot\bm{\Delta}$, where a branch with two linear layers and GELU activation produces the residual correction $\bm{\Delta}$, and a parallel sigmoid gating branch produces $\bm{g}$ to scale the correction term element-wise. The residual formulation provides a direct path for the original visual embedding $\bm{e}_i$, while the gated residual branch learns an additive correction.

\textbf{Remark.} Eq.~\eqref{eq:proto} builds motion prototypes from $\mathsf{T}_{\text{b}}^{\text{emb}}$ rather than from the encoder of RVQ-VAE. The reconstruction loss may emphasize local details over global information~\cite{dong2023peco,ma2025unitok}, whereas $\mathsf{T}_{\text{b}}$ learns to predict motion tokens conditioned on extracted visual observations using Eq.~\eqref{eq:mask}. Fig.~\ref{fig:proto} shows that embeddings from $\mathsf{T}_{\text{b}}^{\text{emb}}$ exhibit clearer separation among reaction categories than those from the RVQ-VAE encoder.

\smallskip
\noindent\emph{2) Stage-2: Observation-Guided Refinement}
\smallskip

MFS rectifies the observation; OGR lets the rectified observation guide the refinement of the coarse motion. The residual Transformer $\mathsf{T}_{\text{r}}$ takes a visual conditioning embedding and first-layer motion tokens as inputs to predict tokens for the remaining quantization layers. OGR combines two complementary designs to improve refinement. First, we combine the rectified and original visual embeddings into the twin conditioning embedding $\bm{c}_i$. The rectified embedding emphasizes cues relevant to reaction generation, while the original embedding complements it with the visual prior captured by the pretrained encoder. Second, during training, $\mathsf{T}_{\text{r}}$ uses coarse tokens $\bar{\bm{s}}_i^{1}$ generated by $\mathsf{T}_{\text{m}}$ from the rectified observation, rather than ground-truth first-layer tokens $\bm{s}_i^{1}$ quantized by the RVQ-VAE as in HERO~\cite{hero}. This design helps $\mathsf{T}_{\text{r}}$ better exploit rectified observations for refinement tailored to $\mathsf{T}_{\text{m}}$'s outputs, while mitigating exposure bias in its coarse-token inputs. The training objective is
\begin{equation}
L_{\text{res}}
=-\mathbb{E}_{i,\,l}\log \mathbb{P}_{\mathsf{T}_{\text{r}}}\Bigl(\bm{s}_i^{l}\Bigm|\bar{\bm{s}}_i^{1},\,\bm{s}_i^{2},\dots,\bm{s}_i^{l-1},\,\bm{c}_{i}\Bigr),
\label{eq:res}
\end{equation}
where the expectation runs over samples and residual layers $l\in\{2,\dots,L+1\}$. For each $l$, $\mathsf{T}_{\text{r}}$ is trained to predict $\bm{s}_i^{l}$ conditioned on the predicted coarse tokens $\bar{\bm{s}}_i^{1}$, the tokens $\bm{s}_i^{2},\dots,\bm{s}_i^{l-1}$ from preceding residual layers, and the twin visual conditioning embedding $\bm{c}_i$.

\textbf{Refinement under enhanced observations.} The rectified embedding $\bm{e}_i'$ reflects improved correspondence between visual observations and reaction motions, while the original embedding $\bm{e}_i$ provides the visual prior captured by the pretrained encoder; both contribute to more effective refinement. Therefore, we employ a Twin-Mixer based on a FiLM-style affine transformation~\cite{perez2018film} to combine the two (Fig.~\ref{fig:arch}~(b)):
\begin{equation}
\bm{c}_{i}=\mathsf{Mix}\bigl(\bm{e}_i',\bm{e}_i\bigr)=\bm{\beta}+(\bm{1}+\bm{\alpha})\odot\bm{e}_i,
\label{eq:cr}
\end{equation}
where $\bm{1}$ is the all-ones vector, and the shift $\bm{\beta}$ and scale adjustment $\bm{\alpha}$ are obtained by splitting $\bm{U}_2\,\phi(\bm{U}_1\bm{e}_i')$ into two equal parts. Here, $\phi$ denotes GELU and $\bm{U}_1$, $\bm{U}_2$ are learnable projection layers, with biases omitted for brevity. This is feature-wise modulation, where original cues stay in $\bm{e}_i$ and rectified guidance enters through $(\bm{\alpha},\bm{\beta})$. $\bm{U}_2$ and its bias are zero-initialized, so training starts from the identity $\bm{c}_{i}=\bm{e}_i$, with the rectified guidance growing from zero.

\textbf{Refinement under $\mathsf{T}_\text{m}$ guidance.} Previous methods~\cite{hero,momask} train teacher-forced residual Transformers conditioned on ground-truth first-layer tokens obtained from RVQ-VAE. Trained this way, $\mathsf{T}_{\text{r}}$ would not be exposed during training to tokens generated by $\mathsf{T}_{\text{m}}$ when conditioned on the rectified observation. This prevents it from adapting its refinement to the resulting errors, creating a train-test gap analogous to exposure bias. We thus start refinement from $\mathsf{T}_{\text{m}}$'s own predictions $\bar{\bm{s}}_i^{1}$. As described in Sec.~\ref{subsec:prel}, these predictions are generated by iteratively filling an all-masked sequence conditioned on the rectified observation. This enables $\mathsf{T}_{\text{r}}$ to learn refinement tailored to the outputs that $\mathsf{T}_{\text{m}}$ generates from rectified observations. The use of generated inputs during training is conceptually related to scheduled sampling~\cite{bengio2015scheduled} and recent strategies in video generation~\cite{zhao2025real,huang2025self}.

During OGR training, only $\mathsf{T}_{\text{r}}$ and $\mathsf{Mix}$ are optimized; $\mathsf{Enc}_{\text{v}}$, $\mathsf{R}$, $\mathsf{T}_{\text{m}}$, and the motion prototypes are frozen. At inference, we encode the video with $\mathsf{Enc}_{\text{v}}$, rectify the embedding with $\mathsf{R}$, generate coarse tokens with $\mathsf{T}_{\text{m}}$, and predict residual tokens with $\mathsf{T}_{\text{r}}$ using visual conditioning from $\mathsf{Mix}$. The resulting tokens are decoded by $\mathsf{Dec}_{\text{m}}$; neither $\mathsf{T}_{\text{b}}$ nor the motion prototypes are needed.

\smallskip

\textbf{Remark.} Compared with HERO~\cite{hero}, our residual stage incorporates two changes: $\mathsf{T}_{\text{r}}$ is conditioned on both rectified and original observations, and it is trained using $\mathsf{T}_{\text{m}}$'s own predictions instead of first-layer tokens directly quantized from reaction motions. Motion prototype feedback in Stage-1 improves observation-reaction correspondence, while Stage-2 combines rectified guidance with the original visual prior and trains $\mathsf{T}_{\text{r}}$ on $\mathsf{T}_{\text{m}}$'s predictions, enabling refinement tailored to outputs generated from rectified observations. HERO~\cite{hero} directly conditions motion generation on frozen visual embeddings without an explicit observation rectification module. Our \texttt{NOUS} explicitly rectifies the correspondence between visual observations and reactions, thus further improving the quality of generated reaction motions.

\section{Experiments}
\label{sec:expe}

\subsection{Experimental Setup}
\label{subsec:es}

\textbf{Benchmark.}
We evaluate on ViMo, introduced by Yu et al.~\cite{hero}, which contains 3,500 videos manually paired with plausible reaction motions from 26 categories. We use the official split of 2,800 training and 700 test pairs. The complete mapping between video and reaction categories, together with the per-category statistics, is provided in Sec.~II of the supplementary material. We report FID, diversity, and multimodality averaged over 20 trials, following the same protocol as Yu et al.~\cite{hero}. FID is the Fréchet distance between the feature distributions of generated and real reactions. Multimodality measures variation among reactions generated from the same video, while diversity measures the overall spread of generated reactions across the test set. For diversity, a score closer to that of real reactions is better. Following Yu et al.~\cite{hero}, we compute FID, diversity, and multimodality using the motion feature extractor of Guo et al.~\cite{guo2022generating} without modification.

\textbf{Compared baselines.} Our primary baseline is HERO~\cite{hero}, which proposed the VHRG task and established its official benchmark. We also compare our \texttt{NOUS} with several open-source motion generation methods adapted to this task by Yu et al., including MDM~\cite{mdm}, MLD~\cite{mld}, T2M-GPT~\cite{t2m-gpt}, BAMM~\cite{bamm}, and MoMask~\cite{momask}. All methods, including ours, are evaluated under the same protocol released by Yu et al., and the results of the compared baselines are taken from the official paper of HERO.

\textbf{Implementation details.}
All experiments are implemented in PyTorch and run on an NVIDIA A6000 GPU. Following Yu et al.~\cite{hero}, we initialize the motion RVQ-VAE with weights pretrained on HumanML3D~\cite{guo2022generating} and fine-tune it on ViMo for 10 epochs. The RVQ-VAE has 6 quantization layers and is fine-tuned with AdamW~\cite{loshchilov2017decoupled} using a learning rate of $2\times10^{-4}$ and a batch size of 256. By default, we use a frozen TC-CLIP~\cite{kim2024leveraging} as the video encoder. The motion and residual Transformers follow the architectures in~\cite{hero} and are trained for 80 and 40 epochs, respectively. Both Transformers are optimized with AdamW~\cite{loshchilov2017decoupled} using a learning rate of $5\times10^{-4}$, a weight decay of $1\times10^{-5}$, and a batch size of 32. The hyperparameters $\mu$ and $\varepsilon$ are set to 30 and 0.4. Following the official implementation of \cite{hero}, we drop the video condition with probability 0.1 during training and apply classifier-free guidance at inference, with guidance scales of 4 and 5 for Stage-1 and Stage-2, respectively.

\subsection{Main Results}
\label{subsec:mr}

\begin{table*}[!tb]
\centering
\caption{\textbf{Quantitative results on the ViMo dataset.} Results are averaged over 20 trials following the evaluation protocol of Yu et al.~\cite{hero}. In the table, ``$\pm$'' denotes the 95\% confidence interval, ``$\rightarrow$'' indicates that results closer to those of real motions (``Real'') are better, and ``$\downarrow$'' and ``$\uparrow$'' indicate that lower and higher values are better, respectively. Bold and underlined values denote the best and second-best results in each column.}
\vspace{-0.2cm}
\small
\setlength{\tabcolsep}{10pt}
\renewcommand{\arraystretch}{1.1}
\begin{tabular}{|l|c|c|c|c|}
\hline
\textbf{Method} & \textbf{Venue} & \textbf{FID} $\downarrow$ & \textbf{Diversity} $\rightarrow$ & \textbf{Multimodality} $\uparrow$ \\
\hline
\hline
Real & -- & -- & 7.954\,$\pm$\,0.074 & -- \\
\hline
MDM \cite{mdm} & ICLR'23 & 1.688\,$\pm$\,0.030 & 7.385\,$\pm$\,0.088 & \textbf{2.117}\,$\pm$\,0.064 \\
MLD \cite{mld} & CVPR'23 & 1.565\,$\pm$\,0.041 & 7.431\,$\pm$\,0.090 & \underline{2.102}\,$\pm$\,0.062 \\
T2M-GPT \cite{t2m-gpt} & CVPR'23 & 1.154\,$\pm$\,0.038 & 7.721\,$\pm$\,0.081 & 1.936\,$\pm$\,0.032 \\
BAMM \cite{bamm} & ECCV'24 & 0.930\,$\pm$\,0.031 & 7.619\,$\pm$\,0.055 & 1.885\,$\pm$\,0.047 \\
MoMask \cite{momask} & CVPR'24 & 0.856\,$\pm$\,0.015 & 7.394\,$\pm$\,0.056 & 1.567\,$\pm$\,0.043 \\
HERO \cite{hero} & ICCV'25 & \underline{0.427}\,$\pm$\,0.014 & \underline{7.801}\,$\pm$\,0.061 & 1.614\,$\pm$\,0.040 \\
\hline
\rowcolor{cyan!8}\texttt{NOUS} (ours) & -- & \textbf{0.328}\,$\pm$\,0.009 & \textbf{7.895}\,$\pm$\,0.065 & 1.648\,$\pm$\,0.038 \\
\hline
\end{tabular}
\label{tab:1}
\vspace{-0.2cm}
\end{table*}

\begin{figure*}[t!]
    \centering
     \includegraphics[width=0.88\linewidth]{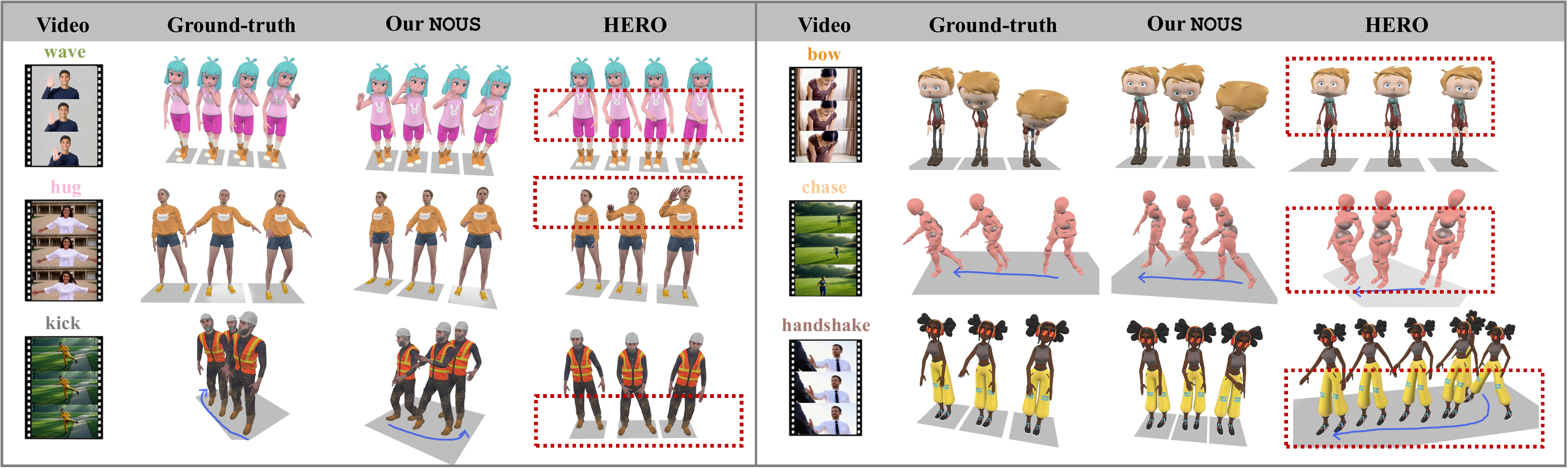}
    \vspace{-0.2cm}
    \caption{\textbf{Qualitative results on ViMo.} Six test videos are arranged in two blocks of three rows, with the interaction label shown to the left of each row. Within each block, the columns show, from left to right, the input frames, ground-truth reaction, and reactions generated by \texttt{NOUS} and HERO~\cite{hero}. Each reaction is visualized through several side-by-side snapshots of the corresponding motion sequence. Blue arrows trace the reactor's ground-plane trajectory and are omitted when planar translation is negligible. Red dashed marks highlight failures in HERO's reactions: a missing hand raise in ``wave'', a weak response in ``kick'', an incomplete ``bow'', and large spurious global translation in ``handshake''.}
    \label{fig:qual}
    \vspace{-0.2cm}
\end{figure*}

\textbf{Quantitative results.} Table~\ref{tab:1} shows that \texttt{NOUS} achieves the lowest FID of 0.328, compared with 0.427 for HERO~\cite{hero}. Its diversity of 7.895 is also closest to the real value of 7.954, whereas HERO reaches 7.801. FID and diversity should be interpreted jointly. MoMask~\cite{momask} has the third-lowest FID overall at 0.856, but its diversity of 7.394 is 0.560 below the real-motion value, indicating a narrower spread in motion feature space. Multimodality should likewise not be interpreted in isolation, because a high value does not necessarily imply high motion fidelity. MDM~\cite{mdm} illustrates this point, attaining the highest multimodality of 2.117 while recording the worst FID of 1.688 and a diversity of 7.385, which is 0.569 below the real value. Although \texttt{NOUS} does not rank first in multimodality, its mean score of 1.648 exceeds those of HERO (1.614) and MoMask (1.567), the two baselines with the lowest FIDs.

\textbf{Qualitative results.} Fig.~\ref{fig:qual} presents qualitative results. The reactions generated by \texttt{NOUS} match the ground truth more closely and better reflect the interaction shown in the input video. By contrast, HERO~\cite{hero} sometimes produces reactions that do not fully match the intended interaction. For example, in wave, the hand barely rises, while in hug, the arm moves toward the head rather than opening outward. In bow, the reactor remains upright instead of completing the bend, and in kick, it shows little backward or lateral movement. HERO also shows a less pronounced running stride in ``chase''; in ``handshake'', it exhibits excessive lateral drift, whereas the ground-truth motion shows little translation. Across these examples, \texttt{NOUS} more closely matches the ground truth in both motion pattern and displacement. Additional visualizations and a user study (all participants gave informed consent; the study collected no personal data) are provided in the supplementary material, with MP4 videos in the multimedia supplement.

\vspace{-0.4cm}
\subsection{Analysis}
\label{subsec:ma}

We examine the following seven aspects of \texttt{NOUS}:
\begin{itemize}
\item The effect of Motion Feedback Steering (MFS) and of each of its components;
\item Alternative feedback designs for observation rectification;
\item The gains from Observation-Guided Refinement (OGR);
\item The roles of Twin-Mixer and $\mathsf{T}_{\text{m}}$'s guidance;
\item Compatibility with other pretrained video encoders;
\item The effect on R-precision in video-motion retrieval;
\item Computational cost.
\end{itemize}

\smallskip
\noindent\emph{1) Analysis of Stage-1: Motion Feedback Steering}
\smallskip

\begin{table}[!tb]
\centering
\caption{\textbf{Ablation study on Motion Feedback Steering (MFS).} To isolate the influence of each component in MFS, all variants are evaluated using Stage-1 outputs only, without Stage-2 refinement. ``w/o MFS'' removes the delta-rectification modulator and $L_{\text{rmc}}$ together, reducing our Stage-1 to that of HERO~\cite{hero}. ``w/ MFS + FT-$\mathsf{Enc}_{\text{v}}$'' retains MFS while unfreezing the pretrained video encoder. The remaining three rows ablate the individual components of the delta-rectification modulator. ``w/o Delta'' removes the residual correction branch and directly adds the gate output to the input, yielding $\mathsf{R}(\bm{x})=\bm{x}+\bm{g}$, ``w/o Linear'' removes the linear projection layers, and ``w/o Gate'' removes the gate $\bm{g}$ that scales the residual correction element-wise.}
\vspace{-0.2cm}
\label{tab:mfs}
\scriptsize
\setlength{\tabcolsep}{3pt}
\renewcommand{\arraystretch}{1.1}
\begin{tabularx}{\linewidth}{|l@{}l|*{3}{>{\centering\arraybackslash}X|}}
\hline
\multicolumn{2}{|l|}{\textbf{Experiments}} & \textbf{FID} $\downarrow$ & \textbf{Diversity} $\rightarrow$ & \textbf{Multimodality} $\uparrow$ \\
\hline
\hline
\multicolumn{2}{|l|}{Real} & -- & 7.954\,$\pm$\,0.074 & -- \\
\hline
\rowcolor{cyan!8} w/ & \,MFS & \textbf{0.366}\,$\pm$\,0.006 & \textbf{8.143}\,$\pm$\,0.099 & \underline{1.198}\,$\pm$\,0.031 \\
w/o & \,MFS & 0.493\,$\pm$\,0.011 & 8.351\,$\pm$\,0.106 & \textbf{1.241}\,$\pm$\,0.032 \\
w/ & \,MFS + FT-$\mathsf{Enc}_{\text{v}}$ & 0.516\,$\pm$\,0.012 & 8.416\,$\pm$\,0.081 &  1.189\,$\pm$\,0.029 \\

\hline
w/o & \,Delta & 0.462\,$\pm$\,0.011 & 8.373\,$\pm$\,0.108 & 1.155\,$\pm$\,0.027 \\
w/o & \,Gate & \underline{0.422}\,$\pm$\,0.008 & \underline{8.320}\,$\pm$\,0.112 & 1.179\,$\pm$\,0.028 \\
w/o & \,Linear & 0.469\,$\pm$\,0.016 & 8.330\,$\pm$\,0.098 & 1.125\,$\pm$\,0.026 \\
\hline
\end{tabularx}
\vspace{-0.1cm}
\end{table}

\textbf{Ablation Study on MFS.} We study the influence of MFS in Table~\ref{tab:mfs}, from which we can draw two observations. First, employing MFS can further improve the quality of reaction motion, lowering FID from 0.493 to 0.366, a reduction of 26\%. Finetuning the pretrained video encoder with MFS increases FID from 0.366 to 0.516, supporting our choice to keep the pretrained video encoder frozen. Second, removing the delta branch, the gate, or the linear layers raises Stage-1 FID from 0.366 to 0.462, 0.422, and 0.469, respectively, supporting the contribution of each component within the full MFS design. MFS slightly lowers multimodality from 1.241 to 1.198, while reducing the absolute deviation of diversity from the real-motion value of 7.954 from 0.397 to 0.189. Fig.~\ref{fig:rd} also shows the corresponding change in the embedding space of visual observations. With MFS, observation embeddings form more distinct clusters by reaction category and exhibit lower off-diagonal cosine similarities, indicating reduced spurious correlations in the visual space.

\textbf{Alternative designs for observation rectification.} Table~\ref{tab:proto} compares our default MFS-MP design with five alternatives: MFS-VP, MFS-MP-VAE, MFS-CE, MFS-MP-LabelFree, and MFS-MP-Instance. MFS-MP and MFS-MP-VAE construct motion prototypes from $\mathsf{T}_{\text{b}}^{\text{emb}}$ and the RVQ-VAE encoder, respectively; MFS-VP constructs visual prototypes from video sequences, while MFS-CE uses reaction-category cross-entropy without motion prototype feedback. MFS-MP-LabelFree builds motion prototypes via unsupervised clustering, whereas MFS-MP-Instance uses individual reaction embeddings instead of mean prototypes for feedback steering. Without OGR, the default MFS-MP achieves the lowest FID of 0.366, compared with 0.430 for MFS-VP, 0.443 for MFS-MP-VAE, and 0.423 for MFS-CE. With OGR, the default again achieves the lowest FID of 0.328, compared with 0.396, 0.397, and 0.386, respectively. Without OGR, MFS-VP lowers FID from 0.493 for the no-rectification baseline in Table~\ref{tab:mfs} to 0.430. Although OGR improves all three metrics for MFS-CE, the resulting scores remain less favorable than those of MFS-MP with OGR. MFS-MP-LabelFree achieves FIDs of 0.384 without OGR and 0.348 with OGR, both lower than HERO's FID of 0.427, indicating that unsupervised prototype construction can still support effective observation rectification. MFS-MP-Instance yields higher FIDs of 0.446 without OGR and 0.414 with OGR. We find that instance-wise motion feedback produces less compact observation clusters within reaction categories than feedback from mean motion prototypes, potentially explaining its higher FIDs both without and with OGR. 
Among the feedback designs in Table~\ref{tab:proto}, MFS-MP achieves the lowest FID both with and without OGR, supporting the use of motion prototypes from $\mathsf{T}_{\text{b}}^{\text{emb}}$ for rectifying extracted visual embeddings.

\begin{table}[t!]
\centering
\caption{\textbf{Comparison of alternative designs for observation rectification.} ``VP'' and ``MP'' denote prototype-based variants that construct feedback anchors from video sequences and reaction motions. Among the motion-based variants, ``MFS-MP-VAE'' builds prototypes using the RVQ-VAE encoder, whereas ``MFS-MP'' uses $\mathsf{T}_{\text{b}}^{\text{emb}}$.
``MFS-MP-LabelFree'' clusters motion embeddings from $\mathsf{T}_{\text{b}}^{\text{emb}}$ using spherical $k$-means ($k\!=\!20$) and averages the embeddings within each cluster to form prototypes without category labels. ``MFS-MP-Instance'' uses instance-wise motion feedback, taking the $\mathsf{T}_{\text{b}}^{\text{emb}}$ representation of each observation's paired reaction as the positive anchor and those of others as negatives.
``MFS-CE'' is equivalent to adding the modulator to HERO's Stage-1 and using reaction-category cross-entropy to rectify visual observations.}
\vspace{-0.2cm}
\label{tab:proto}
\scriptsize
\setlength{\tabcolsep}{5pt}
\renewcommand{\arraystretch}{1}
\begin{tabular}{|l|c|c|c|}
\hline
\textbf{Experiments} & \textbf{FID} $\downarrow$ & \textbf{Diversity} $\rightarrow$ & \textbf{Multimodality} $\uparrow$ \\
\hline
\hline
Real & -- & 7.954\,$\pm$\,0.074 & -- \\
\hline
\rowcolor{cyan!8} MFS-MP & 0.366\,$\pm$\,0.006 & 8.143\,$\pm$\,0.099 & 1.198\,$\pm$\,0.031 \\
MFS-VP & 0.430\,$\pm$\,0.008 & 8.236\,$\pm$\,0.091 &  1.255\,$\pm$\,0.026 \\
MFS-MP-VAE & 0.443\,$\pm$\,0.010 & 8.359\,$\pm$\,0.096 & 1.216\,$\pm$\,0.032 \\
MFS-MP-LabelFree &  0.384\,$\pm$\,0.007 & 8.253\,$\pm$\,0.081 & 1.285\,$\pm$\,0.022 \\
MFS-MP-Instance &  0.446\,$\pm$\,0.012 &  8.227\,$\pm$\,0.085 & 1.443\,$\pm$\,0.033 \\
MFS-CE  & 0.423\,$\pm$\,0.008 & 8.158\,$\pm$\,0.085 & 1.269\,$\pm$\,0.027 \\
\hline
\rowcolor{cyan!8} MFS-MP + OGR & \textbf{0.328}\,$\pm$\,0.009 & 7.895\,$\pm$\,0.065 & \underline{1.648}\,$\pm$\,0.038 \\
MFS-VP + OGR & 0.396\,$\pm$\,0.008 & \underline{7.898}\,$\pm$\,0.071 & 1.532\,$\pm$\,0.035 \\
MFS-MP-VAE + OGR &  0.397\,$\pm$\,0.010 &   8.024\,$\pm$\,0.066 &  1.573\,$\pm$\,0.046 \\
MFS-MP-LabelFree + OGR &  \underline{0.348}\,$\pm$\,0.008 &  \textbf{7.957}\,$\pm$\,0.084 &   1.639\,$\pm$\,0.032 \\
MFS-MP-Instance + OGR & 0.414\,$\pm$\,0.010 & 8.073\,$\pm$\,0.078 &  \textbf{1.710}\,$\pm$\,0.044 \\
MFS-CE + OGR  & 0.386\,$\pm$\,0.011 & 7.874\,$\pm$\,0.081 & 1.517\,$\pm$\,0.034 \\
\hline
\end{tabular}
\end{table}

\begin{figure}[t!]
    \centering
    \includegraphics[width=0.95\linewidth]{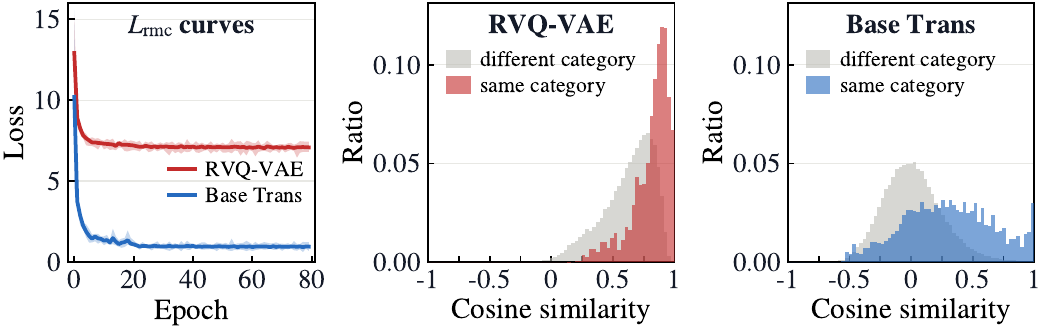}
    \vspace{-0.3cm}
    \caption{\textbf{Comparison of the RVQ-VAE encoder and $\mathsf{T}_{\text{b}}^{\text{emb}}$ (Base Trans) for building motion prototypes.} \textbf{Left:} $L_{\text{rmc}}$ curves during observation rectification with the two prototype sources. \textbf{Middle and right:} distributions of pairwise cosine similarities between the resulting motion embeddings, shown separately for same-category and different-category pairs.}
    \vspace{-0.3cm}
    \label{fig:proto}
\end{figure}

We next examine why $\mathsf{T}_{\text{b}}^{\text{emb}}$ provides better prototypes than the RVQ-VAE encoder. Fig.~\ref{fig:proto} compares their loss curves and cosine similarity distributions. On the left, $L_{\text{rmc}}$ decreases to about~1 with prototypes from $\mathsf{T}_{\mathrm{b}}^{\mathrm{emb}}$, whereas it plateaus near~7 with prototypes from the RVQ-VAE encoder. The histograms show high, substantially overlapping cosine similarity distributions for same- and different-category pairs of RVQ-VAE motion embeddings, indicating limited category separability. In contrast, under $\mathsf{T}_{\text{b}}^{\,\text{emb}}$, different-category similarities cluster near zero, whereas same-category similarities extend toward higher positive values, producing a clearer distinction between the two distributions. As noted earlier, reconstruction loss emphasizes local details but is less effective at capturing global relationships~\cite{dong2023peco,ma2025unitok}. In our setting, RVQ-VAE is trained to reconstruct motions without explicitly enforcing inter-category separation. $\mathsf{T}_{\text{b}}$, by contrast, is conditioned on visual observations through Eq.~\eqref{eq:mask}, thus coupling its motion embeddings with visual cues captured by the pretrained video encoder and making motion representations more discriminative.

\begin{table}[!tb]
\centering
\caption{\textbf{Ablation study on Observation-Guided Refinement (OGR).} ``w/ OGR'' represents using the full Stage-2 of our \texttt{NOUS}, and ``w/o OGR'' disables it. ``w/o Original'' and ``w/o Rectified'' remove the original and rectified observation branches from Twin-Mixer. ``w/o $\mathsf{T}_{\text{m}}$ Guidance'' trains $\mathsf{T}_{\text{r}}$ on first-layer tokens directly quantized by RVQ-VAE instead of $\mathsf{T}_{\text{m}}$'s predictions. ``w/ HERO-RES'' combines ``w/o Rectified'' and ``w/o $\mathsf{T}_{\text{m}}$ Guidance'', replacing our Stage-2 with HERO's residual refinement~\cite{hero} while retaining our Stage-1.}
\vspace{-0.2cm}
\label{tab:ogr}
\scriptsize
\setlength{\tabcolsep}{2pt}
\renewcommand{\arraystretch}{1}
\begin{tabularx}{\linewidth}{|l|*{3}{>{\centering\arraybackslash}X|}}
\hline
\textbf{Experiments} & \textbf{FID} $\downarrow$ & \textbf{Diversity} $\rightarrow$ &\textbf{Multimodality} $\uparrow$ \\
\hline
\hline
Real & -- & 7.954\,$\pm$\,0.074 & -- \\
\hline
\rowcolor{cyan!8} w/\hphantom{o} OGR & \textbf{0.328}\,$\pm$\,0.009 & \textbf{7.895}\,$\pm$\,0.065 & \textbf{1.648}\,$\pm$\,0.038 \\
w/ HERO-RES & 0.361\,$\pm$\,0.008 &  8.088\,$\pm$\,0.061 &  1.037\,$\pm$\,0.030 \\
w/o OGR & 0.366\,$\pm$\,0.006 & 8.143\,$\pm$\,0.099 & 1.198\,$\pm$\,0.031 \\
\hline
w/o $\mathsf{T}_{\text{m}}$ Guidance & 0.359\,$\pm$\,0.009 & 7.809\,$\pm$\,0.071 & 1.567\,$\pm$\,0.033 \\
w/o Original & \underline{0.349}\,$\pm$\,0.012 & 7.832\,$\pm$\,0.083 & 1.596\,$\pm$\,0.030 \\
w/o Rectified & 0.375\,$\pm$\,0.008 & \underline{7.850}\,$\pm$\,0.074 & \underline{1.615}\,$\pm$\,0.031 \\
\hline
\end{tabularx}
\vspace{-0.2cm}
\end{table}

\begin{table}[!tb]
\centering
\caption{\textbf{Compatibility with other pretrained video encoders.} \texttt{NOUS} can be integrated with frozen VideoMAE~\cite{wang2023videomae}, ViCLIP~\cite{wang2023internvid}, and X-CLIP~\cite{ni2022expanding} encoders. For each encoder, ``+ MFS'' adds MFS to HERO-Stage-1, while ``+ OGR'' adds OGR to HERO-Stage-2 while retaining MFS in Stage-1. At both stages, \texttt{NOUS} achieves lower FID and diversity closer to the real-motion value than the corresponding HERO baseline. All experiments follow the TC-CLIP~\cite{kim2024leveraging} training schedule without encoder-specific hyperparameter tuning.}
\vspace{-0.2cm}
\label{tab:enc}
\scriptsize
\setlength{\tabcolsep}{4pt}
\renewcommand{\arraystretch}{1.1}
\begin{tabularx}{\linewidth}{|c|l|>{\hsize=0.9\hsize\centering\arraybackslash}X|>{\hsize=0.95\hsize\centering\arraybackslash}X|>{\hsize=1.15\hsize\centering\arraybackslash}X|}
\hline
\multicolumn{2}{|l|}{\textbf{Experiments}} & \textbf{FID} $\downarrow$ & \textbf{Diversity} $\rightarrow$ & \textbf{Multimodality}~$\uparrow$ \\
\hline
\hline
\multicolumn{2}{|l|}{Real} & -- & 7.954\,$\pm$\,0.074 & -- \\
\hline
& HERO-Stage-1 & 1.209\,$\pm$\,0.030 & 8.381\,$\pm$\,0.071 & 1.370\,$\pm$\,0.058 \\
\rowcolor{cyan!8} & \qquad + MFS & 0.794\,$\pm$\,0.011 & 8.281\,$\pm$\,0.068 & 1.119\,$\pm$\,0.030 \\
\cline{2-5}
& HERO-Stage-2 & 0.961\,$\pm$\,0.023 & 8.325\,$\pm$\,0.076 & 2.555\,$\pm$\,0.070 \\
\rowcolor{cyan!8} \cellcolor{white}\multirow{-4}{*}{\rotatebox[origin=c]{90}{VideoMAE}} & \qquad + OGR & 0.590\,$\pm$\,0.010 & 8.096\,$\pm$\,0.057 & 1.412\,$\pm$\,0.041 \\
\hline
& HERO-Stage-1 & 0.761\,$\pm$\,0.016 & 8.416\,$\pm$\,0.070 & 1.365\,$\pm$\,0.038 \\
\rowcolor{cyan!8} & \qquad + MFS & 0.563\,$\pm$\,0.007 & 8.188\,$\pm$\,0.059 & 1.201\,$\pm$\,0.035 \\
\cline{2-5}
& HERO-Stage-2  & 0.567\,$\pm$\,0.011 & 8.137\,$\pm$\,0.071 & 1.627\,$\pm$\,0.043 \\
\rowcolor{cyan!8} \cellcolor{white}\multirow{-4}{*}{\rotatebox[origin=c]{90}{ViCLIP}} & \qquad + OGR & 0.472\,$\pm$\,0.008 & 7.941\,$\pm$\,0.051 & 1.466\,$\pm$\,0.039 \\
\hline
& HERO-Stage-1 & 0.601\,$\pm$\,0.015 & 8.249\,$\pm$\,0.085 & 1.286\,$\pm$\,0.034 \\
\rowcolor{cyan!8} & \qquad + MFS & 0.415\,$\pm$\,0.006 & 8.183\,$\pm$\,0.052 & 1.120\,$\pm$\,0.029 \\
\cline{2-5}
& HERO-Stage-2 & 0.468\,$\pm$\,0.015 & 8.097\,$\pm$\,0.074 & 1.488\,$\pm$\,0.038 \\
\rowcolor{cyan!8} \cellcolor{white}\multirow{-4}{*}{\rotatebox[origin=c]{90}{X-CLIP}} & \qquad + OGR & 0.315\,$\pm$\,0.007 & 7.946\,$\pm$\,0.045 & 1.402\,$\pm$\,0.036 \\
\hline
\end{tabularx}
\end{table}

\smallskip
\noindent\emph{2) Analysis of Stage-2: Observation-Guided Refinement}
\smallskip

\textbf{Effect of OGR.} We first examine the overall effect of our OGR in Table~\ref{tab:ogr}. OGR improves all three metrics rather than trading one against another. FID falls from 0.366 to 0.328, multimodality rises from 1.198 to 1.648, and diversity moves from 8.143 to 7.895, that is toward real motion rather than further above it. The absolute deviation of the diversity score from the real-motion value of 7.954 decreases from 0.189 to 0.059. The increase in multimodality indicates greater variation among reactions generated from the same observation. Pairing HERO's Stage-2 with our Stage-1 yields an FID of 0.361, a diversity of 8.088, and a multimodality of 1.037, underperforming OGR on all three metrics. 

\textbf{Analysis of Twin-Mixer and $\mathsf{T}_{\text{m}}$ guidance.} With $\mathsf{T}_{\text{m}}$ guidance, the results in Table~\ref{tab:ogr} support retaining both the original and rectified visual embeddings in Twin-Mixer. Removing either one raises FID, to 0.349 when only the rectified observation is kept and to 0.375 when only the original is kept. This is consistent with our design: the rectified embedding provides reaction-relevant guidance, while the original embedding supplies visual cues from the pretrained video encoder. With Twin-Mixer retained, replacing ground-truth first-layer tokens~\cite{hero} with $\mathsf{T}_{\text{m}}$'s predictions during training lowers FID from 0.359 to 0.328. With only the original embedding, however, the same change raises FID from 0.361 to 0.375. These results suggest that combining the original and rectified visual observations as conditioning with $\mathsf{T}_{\text{m}}$ guidance helps $\mathsf{T}_{\text{r}}$ learn refinement tailored to outputs generated from rectified observations.

\begin{figure}[t!]
    \centering
\includegraphics[width=1\linewidth]{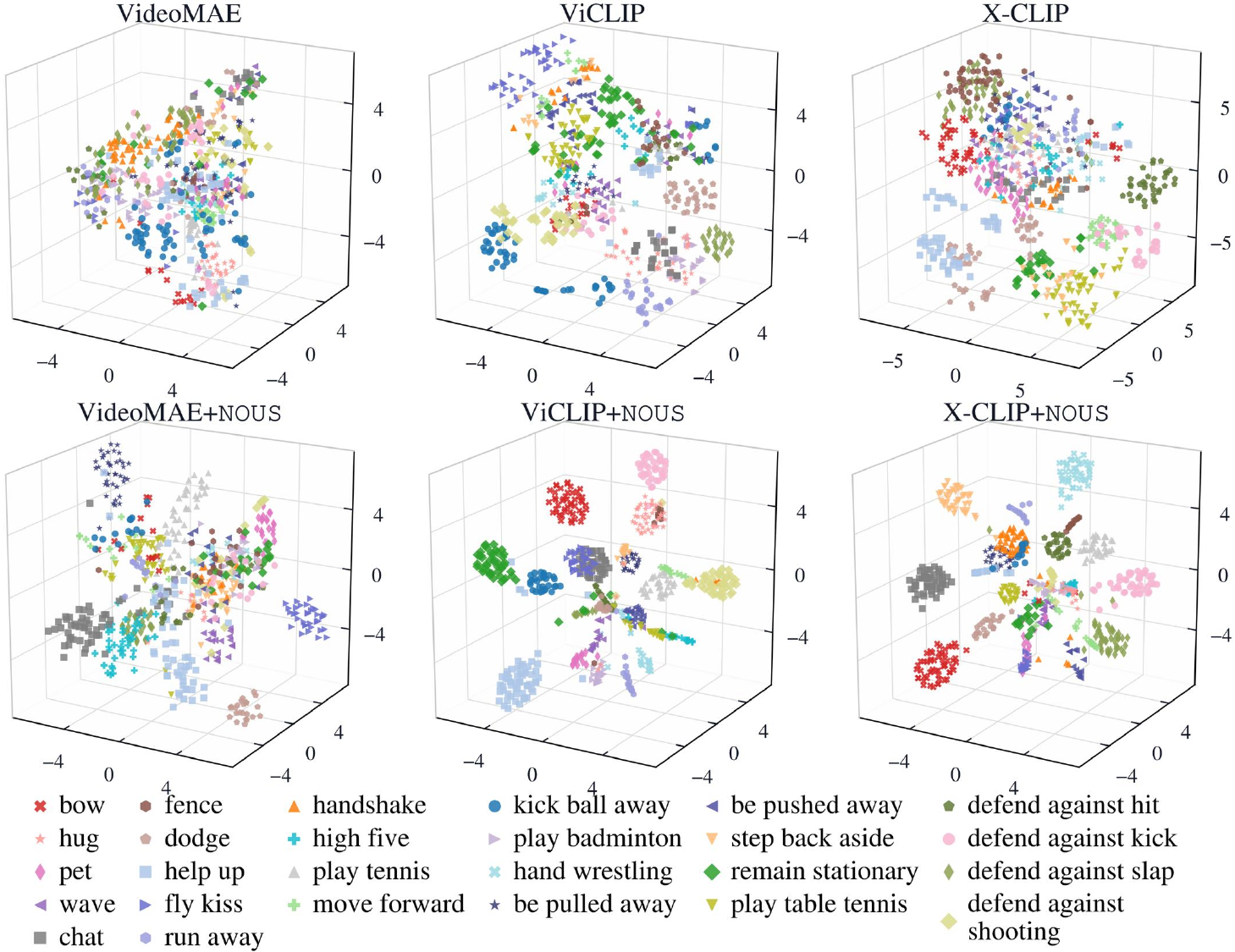}
\vspace{-0.6cm}
\caption{\textbf{t-SNE visualizations of observation embeddings before and after rectification by \texttt{NOUS}.} From left to right, the columns correspond to frozen VideoMAE~\cite{wang2023videomae}, ViCLIP~\cite{wang2023internvid}, and X-CLIP~\cite{ni2022expanding} encoders. The top and bottom rows show the extracted visual embeddings before and after rectification, respectively. Each point represents a test video and is colored by its ground-truth reaction category. Across all three encoders, the rectified visual observations form more clearly separated clusters by reaction category.}
\label{fig:compatibility}
\end{figure}

\smallskip
\noindent\emph{3) Compatibility, R-precision, and Efficiency}
\smallskip

\textbf{Compatibility.} We further analyze whether our \texttt{NOUS} framework can be applied to other pretrained video encoders. Table~\ref{tab:enc} compares \texttt{NOUS} with HERO at both stages using three frozen video encoders. VideoMAE~\cite{wang2023videomae} is trained with a masked reconstruction objective, ViCLIP~\cite{wang2023internvid} is trained with video-text contrastive learning, and X-CLIP~\cite{ni2022expanding} extends image-text contrastive training to video. All three encoders follow the TC-CLIP~\cite{kim2024leveraging} training schedule without encoder-specific hyperparameter tuning. At Stage-1, MFS lowers FID relative to HERO from 1.209 to 0.794 on VideoMAE, from 0.761 to 0.563 on ViCLIP, and from 0.601 to 0.415 on X-CLIP. At Stage-2, \texttt{NOUS} with both MFS and OGR lowers FID relative to HERO from 0.961 to 0.590 on VideoMAE, from 0.567 to 0.472 on ViCLIP, and from 0.468 to 0.315 on X-CLIP. Diversity is also closer to the real-motion value of 7.954 in all six comparisons. Multimodality is lower than HERO's at both stages, although it increases from our Stage-1 to Stage-2.  Fig.~\ref{fig:compatibility} compares the embedding space with and without our \texttt{NOUS} on these three encoders. Without it, the extracted visual embeddings are not organized by reaction category on any of the three encoders, and videos calling for different reactions sit close in the visual space. With \texttt{NOUS}, extracted visual observations from different reaction categories become more clearly separated, while those within the same category form more coherent clusters. These results demonstrate the good compatibility of our \texttt{NOUS} with different pretrained video encoders.

\begin{figure}[!tb]
\centering
\includegraphics[width=0.95\linewidth]{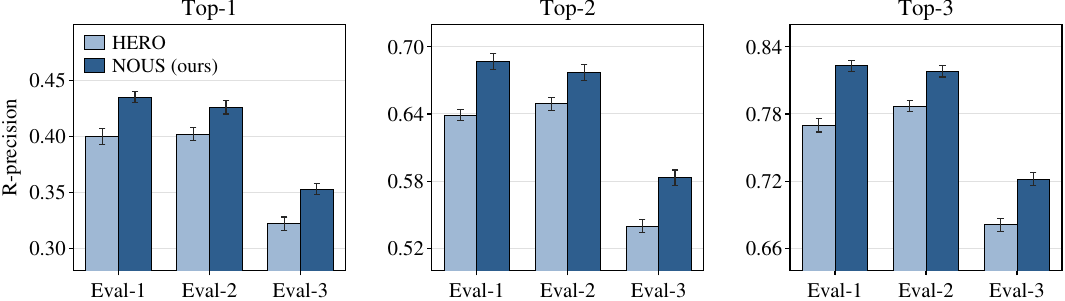}
\vspace{-0.3cm}
\caption{\textbf{R-precision under three video-motion evaluators on the ViMo test set.} The evaluators use the motion feature extractor of Guo et al.~\cite{guo2022generating} and differ only in the video encoder: TC-CLIP~\cite{kim2024leveraging} for Evaluator-1, X-CLIP~\cite{ni2022expanding} for Evaluator-2, and VideoMAE~\cite{wang2023videomae} for Evaluator-3. \texttt{NOUS} achieves higher Top-1, Top-2, and Top-3 R-precision than HERO~\cite{hero} under all three evaluators. Error bars are 95\% confidence intervals over 20 trials.}
\label{fig:rprecision}
\vspace{-0.2cm}
\end{figure}

\textbf{R-precision for video-motion retrieval.} FID compares the distribution of generated motions with that of real ones, but does not measure whether each reaction matches the video that conditioned it. HERO~\cite{hero} does not evaluate this on ViMo due to the lack of a retrieval evaluator. We therefore adapt the retrieval evaluation framework of Guo et al.~\cite{guo2022generating} to VHRG, train three video-motion evaluators, and report R-precision in Fig.~\ref{fig:rprecision}. The results indicate that our \texttt{NOUS} is consistently better than HERO~\cite{hero} under all three evaluators. On Evaluator-1, it achieves Top-1, Top-2, and Top-3 R-precision scores of 0.435, 0.687, and 0.823, respectively, outperforming HERO’s corresponding scores of 0.400, 0.639, and 0.770. On Evaluator-2, it reaches 0.426, 0.677, and 0.818, against 0.402, 0.649, and 0.787; and on Evaluator-3, it reaches 0.353, 0.583, and 0.722, against 0.322, 0.540, and 0.681. The \texttt{NOUS} model evaluated here uses TC-CLIP for video conditioning, whereas Evaluator-2 and Evaluator-3 use X-CLIP and VideoMAE, respectively. The gains persist under these three evaluators, providing further evidence that \texttt{NOUS} improves the correspondence between videos and reaction motions.

\begin{table}[!tb]
\centering
\caption{\textbf{Efficiency analysis.} $\mathsf{T}_{\text{m}}$ and $\mathsf{T}_{\text{r}}$ denote the motion and residual Transformers adopted from HERO~\cite{hero}; MFS and OGR are our key components. Check marks indicate which components are enabled, and the shaded row corresponds to the full \texttt{NOUS}. Together, MFS and OGR add 2.1\,M parameters, 1.0\,G FLOPs, and 0.2\,ms of runtime per sample. FLOPs are computed for a motion with a length of 200 frames; runtime is averaged over 100 runs on an NVIDIA A6000 GPU with a batch size of 32 and reported per sample.}
\vspace{-0.2cm}
\label{tab:eff}
\scriptsize
\setlength{\tabcolsep}{9.2pt}
\renewcommand{\arraystretch}{1}
\begin{tabular}{|c|c|c|c|c|c|c|}
\hline
$\mathsf{T}_{\text{m}}$ & $\mathsf{T}_{\text{r}}$ & \textbf{MFS} & \textbf{OGR} & \textbf{Params} & \textbf{FLOPs} & \textbf{Runtime} \\
\hline
\hline
\checkmark & & & & 16.0\,M & 23.9\,G & 3.1\,ms \\
\checkmark & \checkmark & & & 31.2\,M & 36.1\,G & 5.0\,ms \\
\hline
\checkmark & \checkmark & \checkmark & & 32.5\,M & 37.0\,G & 5.2\,ms \\
\rowcolor{cyan!8} \checkmark & \checkmark & \checkmark & \checkmark & 33.3\,M & 37.1\,G & 5.2\,ms \\
\hline
\end{tabular}
\vspace{-0.3cm}
\end{table}

\textbf{Computational cost.} We also analyze the additional computational overhead incurred by \texttt{NOUS} in practical deployment. Table~\ref{tab:eff} separates the cost of the two stages. It shows that MFS adds 1.3\,M parameters, 0.9\,G FLOPs, and 0.2\,ms per sample, while OGR adds 0.8\,M parameters and 0.1\,G FLOPs, with runtime remaining at 5.2\,ms at the reported precision. Together, MFS and OGR add 2.1\,M parameters and 1.0\,G FLOPs, corresponding to increases of approximately 7\% and 3\% over the baseline comprising $\mathsf{T}_{\text{m}}$ and $\mathsf{T}_{\text{r}}$. The average runtime per sample increases from 5.0\,ms to 5.2\,ms, an increase of 4\%. These results indicate that \texttt{NOUS} achieves improvements with little additional computation during inference. 

\vspace{-0.2cm}
\section{Conclusion and Limitations}
We identify relational distortion in video-driven human reaction generation: a misaligned correspondence between videos and reactions, whereby videos lying close in the visual embedding space may call for entirely different reactions. To address this issue, we propose \texttt{NOUS}, a framework based on observation-reaction mutual steering. MFS trains a lightweight delta-rectification modulator with motion prototype feedback to rectify visual observations. OGR then uses the original and rectified observations to guide refinement tailored to the coarse outputs generated in Stage-1. Experiments on ViMo show that \texttt{NOUS} consistently improves FID and diversity across four pretrained video encoders, with little additional computation during inference.

Video-driven human reaction generation is still at an early stage. Although it carries great potential value for interactive multimedia systems and embodied agents, its data resources remain scarce, as paired video-motion data are costly to collect. ViMo remains the only publicly released benchmark. Text-motion data, by contrast, is more abundant. In future work, we will explore transfer learning from such text-motion pairs to further improve video-driven human reaction generation.

\vspace{-0.3cm}
\bibliographystyle{IEEEtran}

\bibliography{ref}   

@IEEEtranBSTCTL{BSTcontrol,
  CTLuse_forced_etal       = "yes",
  CTLmax_names_forced_etal = "4",
  CTLnames_show_etal       = "3"
}

@String(AAAI  = {AAAI})

@inproceedings{mdm,
  title={Human Motion Diffusion Model},
  author={Tevet, Guy and Raab, Sigal and Gordon, Brian and Shafir, Yoni and Cohen-Or, Daniel and Bermano, Amit Haim},
  booktitle={Proceedings of the International Conference on Learning Representations},
  year={2023}
}

@inproceedings{mld,
  title={Executing your commands via motion diffusion in latent space},
  author={Chen, Xin and Jiang, Biao and Liu, Wen and Huang, Zilong and Fu, Bin and Chen, Tao and Yu, Gang},
  booktitle={Proceedings of the IEEE/CVF Conference on Computer Vision and Pattern Recognition},
  pages={18000--18010},
  year={2023}
}

@inproceedings{bamm,
  title={{BAMM}: Bidirectional autoregressive motion model},
  author={Pinyoanuntapong, Ekkasit and Saleem, Muhammad Usama and Wang, Pu and Lee, Minwoo and Das, Srijan and Chen, Chen},
  booktitle={Proceedings of the European Conference on Computer Vision},
  pages={172--190},
  year={2024},
  organization={Springer}
}

@inproceedings{t2m-gpt,
  title={Generating human motion from textual descriptions with discrete representations},
  author={Zhang, Jianrong and Zhang, Yangsong and Cun, Xiaodong and Zhang, Yong and Zhao, Hongwei and Lu, Hongtao and Shen, Xi and Shan, Ying},
  booktitle={Proceedings of the IEEE/CVF Conference on Computer Vision and Pattern Recognition},
  pages={14730--14740},
  year={2023}
}

@inproceedings{momask,
  title={Momask: Generative masked modeling of 3d human motions},
  author={Guo, Chuan and Mu, Yuxuan and Javed, Muhammad Gohar and Wang, Sen and Cheng, Li},
  booktitle={Proceedings of the IEEE/CVF Conference on Computer Vision and Pattern Recognition},
  pages={1900--1910},
  year={2024}
}

@inproceedings{hero,
  title={Hero: Human reaction generation from videos},
  author={Yu, Chengjun and Zhai, Wei and Yang, Yuhang and Cao, Yang and Zha, Zheng-Jun},
  booktitle={Proceedings of the IEEE/CVF International Conference on Computer Vision},
  pages={10262--10274},
  year={2025},
  organization={IEEE}
}

@inproceedings{athanasiou2022teach,
  title={Teach: Temporal action composition for 3d humans},
  author={Athanasiou, Nikos and Petrovich, Mathis and Black, Michael J and Varol, G{\"u}l},
  booktitle={Proceedings of the International Conference on 3D Vision},
  pages={414--423},
  year={2022},
  organization={IEEE}
}

@inproceedings{cervantes2022implicit,
  title={Implicit neural representations for variable length human motion generation},
  author={Cervantes, Pablo and Sekikawa, Yusuke and Sato, Ikuro and Shinoda, Koichi},
  booktitle={Proceedings of the European Conference on Computer Vision},
  pages={356--372},
  year={2022},
  organization={Springer}
}

@inproceedings{guo2020action2motion,
  title={Action2motion: Conditioned generation of 3d human motions},
  author={Guo, Chuan and Zuo, Xinxin and Wang, Sen and Zou, Shihao and Sun, Qingyao and Deng, Annan and Gong, Minglun and Cheng, Li},
  booktitle={Proceedings of the 28th ACM International Conference on Multimedia},
  pages={2021--2029},
  year={2020}
}

@inproceedings{petrovich2021action,
  title={Action-conditioned 3d human motion synthesis with transformer vae},
  author={Petrovich, Mathis and Black, Michael J and Varol, G{\"u}l},
  booktitle={Proceedings of the IEEE/CVF International Conference on Computer Vision},
  pages={10985--10995},
  year={2021}
}

@inproceedings{xu2023actformer,
  title={Actformer: A gan-based transformer towards general action-conditioned 3d human motion generation},
  author={Xu, Liang and Song, Ziyang and Wang, Dongliang and Su, Jing and Fang, Zhicheng and Ding, Chenjing and Gan, Weihao and Yan, Yichao and Jin, Xin and Yang, Xiaokang and others},
  booktitle={Proceedings of the IEEE/CVF International Conference on Computer Vision},
  pages={2228--2238},
  year={2023}
}

@inproceedings{guo2022generating,
  title={Generating diverse and natural 3d human motions from text},
  author={Guo, Chuan and Zou, Shihao and Zuo, Xinxin and Wang, Sen and Ji, Wei and Li, Xingyu and Cheng, Li},
  booktitle={Proceedings of the IEEE/CVF Conference on Computer Vision and Pattern Recognition},
  pages={5152--5161},
  year={2022}
}

@inproceedings{guo2022tm2t,
  title={{TM2T}: Stochastic and tokenized modeling for the reciprocal generation of 3d human motions and texts},
  author={Guo, Chuan and Zuo, Xinxin and Wang, Sen and Cheng, Li},
  booktitle={Proceedings of the European Conference on Computer Vision},
  pages={580--597},
  year={2022},
  organization={Springer}
}

@inproceedings{petrovich2022temos,
  title={Temos: Generating diverse human motions from textual descriptions},
  author={Petrovich, Mathis and Black, Michael J and Varol, G{\"u}l},
  booktitle={Proceedings of the European Conference on Computer Vision},
  pages={480--497},
  year={2022},
  organization={Springer}
}

@inproceedings{gong2023tm2d,
  title={Tm2d: Bimodality driven 3d dance generation via music-text integration},
  author={Gong, Kehong and Lian, Dongze and Chang, Heng and Guo, Chuan and Jiang, Zihang and Zuo, Xinxin and Mi, Michael Bi and Wang, Xinchao},
  booktitle={Proceedings of the IEEE/CVF International Conference on Computer Vision},
  pages={9942--9952},
  year={2023}
}

@inproceedings{tseng2023edge,
  title={Edge: Editable dance generation from music},
  author={Tseng, Jonathan and Castellon, Rodrigo and Liu, Karen},
  booktitle={Proceedings of the IEEE/CVF Conference on Computer Vision and Pattern Recognition},
  pages={448--458},
  year={2023}
}

@inproceedings{wang2020learning,
  title={Learning diverse stochastic human-action generators by learning smooth latent transitions},
  author={Wang, Zhenyi and Yu, Ping and Zhao, Yang and Zhang, Ruiyi and Zhou, Yufan and Yuan, Junsong and Chen, Changyou},
  booktitle={Proceedings of the AAAI Conference on Artificial Intelligence},
  volume={34},
  number={07},
  pages={12281--12288},
  year={2020}
}

@inproceedings{yan2019convolutional,
  title={Convolutional sequence generation for skeleton-based action synthesis},
  author={Yan, Sijie and Li, Zhizhong and Xiong, Yuanjun and Yan, Huahan and Lin, Dahua},
  booktitle={Proceedings of the IEEE/CVF International Conference on Computer Vision},
  pages={4394--4402},
  year={2019}
}

@article{goodfellow2014generative,
  title={Generative adversarial nets},
  author={Goodfellow, Ian J and Pouget-Abadie, Jean and Mirza, Mehdi and Xu, Bing and Warde-Farley, David and Ozair, Sherjil and Courville, Aaron and Bengio, Yoshua},
  journal={Proceedings of the Advances in Neural Information Processing Systems},
  volume={27},
  year={2014}
}

@inproceedings{kingma2013auto,
  title={Auto-Encoding Variational {Bayes}},
  author={Kingma, Diederik P and Welling, Max},
  booktitle={Proceedings of the International Conference on Learning Representations},
  year={2014}
}

@article{ho2020denoising,
  title={Denoising diffusion probabilistic models},
  author={Ho, Jonathan and Jain, Ajay and Abbeel, Pieter},
  journal={Proceedings of the Advances in Neural Information Processing Systems},
  volume={33},
  pages={6840--6851},
  year={2020}
}

@inproceedings{sohl2015deep,
  title={Deep unsupervised learning using nonequilibrium thermodynamics},
  author={Sohl-Dickstein, Jascha and Weiss, Eric and Maheswaranathan, Niru and Ganguli, Surya},
  booktitle={Proceedings of the International Conference on Machine Learning},
  pages={2256--2265},
  year={2015},
  organization={pmlr}
}

@inproceedings{
gohar2024intermask,
title={InterMask: 3D Human Interaction Generation via Collaborative Masked Modeling},
author={Muhammad Gohar Javed and Chuan Guo and Li Cheng and Xingyu Li},
booktitle={Proceedings of the International Conference on Learning Representations},
year={2025},
url={https://openreview.net/forum?id=ZAyuwJYN8N}
}

@inproceedings{pinyoanuntapong2024mmm,
  title={{MMM}: Generative masked motion model},
  author={Pinyoanuntapong, Ekkasit and Wang, Pu and Lee, Minwoo and Chen, Chen},
  booktitle={Proceedings of the IEEE/CVF Conference on Computer Vision and Pattern Recognition},
  pages={1546--1555},
  year={2024}
}

@inproceedings{dai2024motionlcm,
  title={{MotionLCM}: Real-time controllable motion generation via latent consistency model},
  author={Dai, Wenxun and Chen, Ling-Hao and Wang, Jingbo and Liu, Jinpeng and Dai, Bo and Tang, Yansong},
  booktitle={Proceedings of the European Conference on Computer Vision},
  pages={390--408},
  year={2024},
  organization={Springer}
}

@inproceedings{huang2024stablemofusion,
  title={Stablemofusion: Towards robust and efficient diffusion-based motion generation framework},
  author={Huang, Yiheng and Yang, Hui and Luo, Chuanchen and Wang, Yuxi and Xu, Shibiao and Zhang, Zhaoxiang and Zhang, Man and Peng, Junran},
  booktitle={Proceedings of the 32nd ACM International Conference on Multimedia},
  pages={224--232},
  year={2024}
}

@inproceedings{kim2023flame,
  title={Flame: Free-form language-based motion synthesis \& editing},
  author={Kim, Jihoon and Kim, Jiseob and Choi, Sungjoon},
  booktitle={Proceedings of the AAAI Conference on Artificial Intelligence},
  volume={37},
  number={7},
  pages={8255--8263},
  year={2023}
}

@inproceedings{kong2023priority,
  title={Priority-centric human motion generation in discrete latent space},
  author={Kong, Hanyang and Gong, Kehong and Lian, Dongze and Mi, Michael Bi and Wang, Xinchao},
  booktitle={Proceedings of the IEEE/CVF International Conference on Computer Vision},
  pages={14806--14816},
  year={2023}
}

@inproceedings{shafir2023human,
  title={Human motion diffusion as a generative prior},
  author={Shafir, Yonatan and Tevet, Guy and Kapon, Roy and Bermano, Amit},
  booktitle={Proceedings of the International Conference on Learning Representations},
  year={2024}
}

@inproceedings{yuan2023physdiff,
  title={Physdiff: Physics-guided human motion diffusion model},
  author={Yuan, Ye and Song, Jiaming and Iqbal, Umar and Vahdat, Arash and Kautz, Jan},
  booktitle={Proceedings of the IEEE/CVF International Conference on Computer Vision},
  pages={16010--16021},
  year={2023}
}

@inproceedings{zhou2024emdm,
  title={Emdm: Efficient motion diffusion model for fast and high-quality motion generation},
  author={Zhou, Wenyang and Dou, Zhiyang and Cao, Zeyu and Liao, Zhouyingcheng and Wang, Jingbo and Wang, Wenjia and Liu, Yuan and Komura, Taku and Wang, Wenping and Liu, Lingjie},
  booktitle={Proceedings of the European Conference on Computer Vision},
  pages={18--38},
  year={2024},
  organization={Springer}
}

@article{jiang2023motiongpt,
  title={Motiongpt: Human motion as a foreign language},
  author={Jiang, Biao and Chen, Xin and Liu, Wen and Yu, Jingyi and Yu, Gang and Chen, Tao},
  journal={Proceedings of the Advances in Neural Information Processing Systems},
  volume={36},
  pages={20067--20079},
  year={2023}
}

@inproceedings{zhang2024motiongpt,
  title={Motiongpt: Finetuned llms are general-purpose motion generators},
  author={Zhang, Yaqi and Huang, Di and Liu, Bin and Tang, Shixiang and Lu, Yan and Chen, Lu and Bai, Lei and Chu, Qi and Yu, Nenghai and Ouyang, Wanli},
  booktitle={Proceedings of the AAAI Conference on Artificial Intelligence},
  volume={38},
  number={7},
  pages={7368--7376},
  year={2024}
}

@inproceedings{zhong2023attt2m,
  title={Attt2m: Text-driven human motion generation with multi-perspective attention mechanism},
  author={Zhong, Chongyang and Hu, Lei and Zhang, Zihao and Xia, Shihong},
  booktitle={Proceedings of the IEEE/CVF International Conference on Computer Vision},
  pages={509--519},
  year={2023}
}

@article{chopin2023interaction,
  title={Interaction transformer for human reaction generation},
  author={Chopin, Baptiste and Tang, Hao and Otberdout, Naima and Daoudi, Mohamed and Sebe, Nicu},
  journal={IEEE Transactions on Multimedia},
  volume={25},
  pages={8842--8854},
  year={2023},
  publisher={IEEE}
}

@inproceedings{ghosh2024remos,
  title={Remos: 3d motion-conditioned reaction synthesis for two-person interactions},
  author={Ghosh, Anindita and Dabral, Rishabh and Golyanik, Vladislav and Theobalt, Christian and Slusallek, Philipp},
  booktitle={Proceedings of the European Conference on Computer Vision},
  pages={418--437},
  year={2024},
  organization={Springer}
}

@article{liu2023interactive,
  title={Interactive humanoid: Online full-body motion reaction synthesis with social affordance canonicalization and forecasting},
  author={Liu, Yunze and Chen, Changxi and Yi, Li},
  journal={arXiv preprint arXiv:2312.08983},
  year={2023}
}

@inproceedings{liu2024physreaction,
  title={PhysReaction: Physically plausible real-time humanoid reaction synthesis via forward dynamics guided 4d imitation},
  author={Liu, Yunze and Chen, Changxi and Ding, Chenjing and Yi, Li},
  booktitle={Proceedings of the 32nd ACM International Conference on Multimedia},
  pages={3771--3780},
  year={2024}
}

@inproceedings{xu2024regennet,
  title={{ReGenNet}: Towards human action-reaction synthesis},
  author={Xu, Liang and Zhou, Yizhou and Yan, Yichao and Jin, Xin and Zhu, Wenhan and Rao, Fengyun and Yang, Xiaokang and Zeng, Wenjun},
  booktitle={Proceedings of the IEEE/CVF Conference on Computer Vision and Pattern Recognition},
  pages={1759--1769},
  year={2024}
}

@article{zeghidour2021soundstream,
  title={{SoundStream}: An end-to-end neural audio codec},
  author={Zeghidour, Neil and Luebs, Alejandro and Omran, Ahmed and Skoglund, Jan and Tagliasacchi, Marco},
  journal={IEEE/ACM Transactions on Audio, Speech, and Language Processing},
  volume={30},
  pages={495--507},
  year={2021},
  publisher={IEEE}
}

@inproceedings{chang2022maskgit,
  title={{MaskGIT}: Masked generative image transformer},
  author={Chang, Huiwen and Zhang, Han and Jiang, Lu and Liu, Ce and Freeman, William T},
  booktitle={Proceedings of the IEEE/CVF Conference on Computer Vision and Pattern Recognition},
  pages={11315--11325},
  year={2022}
}

@inproceedings{wang2025timotion,
  title={TIMotion: Temporal and Interactive Framework for Efficient Human-Human Motion Generation},
  author={Wang, Yabiao and Wang, Shuo and Zhang, Jiangning and Fan, Ke and Wu, Jiafu and Xue, Zhucun and Liu, Yong},
  booktitle={Proceedings of the Computer Vision and Pattern Recognition Conference},
  pages={7169--7178},
  year={2025}
}

@inproceedings{
tan2025think,
title={Think Then React: Towards Unconstrained Action-to-Reaction Motion Generation},
author={Wenhui Tan and Boyuan Li and Chuhao Jin and Wenbing Huang and Xiting Wang and Ruihua Song},
booktitle={Proceedings of the International Conference on Learning Representations},
year={2025},
url={https://openreview.net/forum?id=UxzKcIZedp}
}

@inproceedings{dong2023peco,
  title={Peco: Perceptual codebook for bert pre-training of vision transformers},
  author={Dong, Xiaoyi and Bao, Jianmin and Zhang, Ting and Chen, Dongdong and Zhang, Weiming and Yuan, Lu and Chen, Dong and Wen, Fang and Yu, Nenghai and Guo, Baining},
  booktitle={Proceedings of the AAAI Conference on Artificial Intelligence},
  volume={37},
  number={1},
  pages={552--560},
  year={2023}
}

@inproceedings{loshchilov2017decoupled,
  title={Decoupled Weight Decay Regularization},
  author={Loshchilov, Ilya and Hutter, Frank},
  booktitle={Proceedings of the International Conference on Learning Representations},
  year={2019}
}

@inproceedings{Zhu_2025_ICCV,
  title={Distilling parallel gradients for fast ODE solvers of diffusion models},
  author={Zhu, Beier and Wang, Ruoyu and Zhao, Tong and Zhang, Hanwang and Zhang, Chi},
  booktitle={Proceedings of the IEEE/CVF International Conference on Computer Vision},
  pages={19557--19566},
  year={2025},
  organization={IEEE}
}

@inproceedings{kim2024leveraging,
  title={Leveraging temporal contextualization for video action recognition},
  author={Kim, Minji and Han, Dongyoon and Kim, Taekyung and Han, Bohyung},
  booktitle={Proceedings of the European Conference on Computer Vision},
  pages={74--91},
  year={2024},
  organization={Springer}
}

@article{zhao2025real,
  title={Real-Time Motion-Controllable Autoregressive Video Diffusion},
  author={Zhao, Kesen and Shi, Jiaxin and Zhu, Beier and Zhou, Junbao and Shen, Xiaolong and Zhou, Yuan and Sun, Qianru and Zhang, Hanwang},
  journal={arXiv preprint arXiv:2510.08131},
  year={2025}
}

@inproceedings{wang2023videomae,
  title={Videomae v2: Scaling video masked autoencoders with dual masking},
  author={Wang, Limin and Huang, Bingkun and Zhao, Zhiyu and Tong, Zhan and He, Yinan and Wang, Yi and Wang, Yali and Qiao, Yu},
  booktitle={Proceedings of the IEEE/CVF Conference on Computer Vision and Pattern Recognition},
  pages={14549--14560},
  year={2023}
}

@inproceedings{wang2023internvid,
  title={InternVid: A Large-scale Video-Text Dataset for Multimodal Understanding and Generation},
  author={Wang, Yi and He, Yinan and Li, Yizhuo and Li, Kunchang and Yu, Jiashuo and Ma, Xin and Li, Xinhao and Chen, Guo and Chen, Xinyuan and Wang, Yaohui and others},
  booktitle={Proceedings of the International Conference on Learning Representations},
  year={2024}
}

@inproceedings{ni2022expanding,
  title={Expanding language-image pretrained models for general video recognition},
  author={Ni, Bolin and Peng, Houwen and Chen, Minghao and Zhang, Songyang and Meng, Gaofeng and Fu, Jianlong and Xiang, Shiming and Ling, Haibin},
  booktitle={Proceedings of the European Conference on Computer Vision},
  pages={1--18},
  year={2022},
  organization={Springer}
}

@inproceedings{bengio2015scheduled,
  title     = {Scheduled Sampling for Sequence Prediction with Recurrent Neural Networks},
  author    = {Bengio, Samy and Vinyals, Oriol and Jaitly, Navdeep and Shazeer, Noam},
  booktitle = {Proceedings of the Advances in Neural Information Processing Systems},
  year      = {2015}
}

@inproceedings{perez2018film,
  title     = {FiLM: Visual Reasoning with a General Conditioning Layer},
  author    = {Perez, Ethan and Strub, Florian and De Vries, Harm and Dumoulin, Vincent and Courville, Aaron},
  booktitle = {Proceedings of the AAAI Conference on Artificial Intelligence},
  year      = {2018}
}

@article{sun2020deepdance,
  title={Deepdance: music-to-dance motion choreography with adversarial learning},
  author={Sun, Guofei and Wong, Yongkang and Cheng, Zhiyong and Kankanhalli, Mohan S and Geng, Weidong and Li, Xiangdong},
  journal={IEEE Transactions on Multimedia},
  volume={23},
  pages={497--509},
  year={2020},
  publisher={IEEE}
}

@article{qi2024emotiongesture,
  title={Emotiongesture: Audio-driven diverse emotional co-speech 3d gesture generation},
  author={Qi, Xingqun and Liu, Chen and Li, Lincheng and Hou, Jie and Xin, Haoran and Yu, Xin},
  journal={IEEE Transactions on Multimedia},
  volume={26},
  pages={10420--10430},
  year={2024},
  publisher={IEEE}
}

@article{wang2024crossmodal,
  title={Cross-modal quantization for co-speech gesture generation},
  author={Wang, Zheng and Zhang, Wei and Ye, Long and Zeng, Dan and Mei, Tao},
  journal={IEEE Transactions on Multimedia},
  volume={26},
  pages={10251--10263},
  year={2024},
  publisher={IEEE}
}

@article{shi2023multi,
  title={Multi-semantics aggregation network based on the dynamic-attention mechanism for 3d human motion prediction},
  author={Shi, Junyu and Zhong, Jianqi and Cao, Wenming},
  journal={IEEE Transactions on Multimedia},
  volume={26},
  pages={5194--5206},
  year={2023},
  publisher={IEEE}
}

@article{yu2024divdiff,
  title={Divdiff: A conditional diffusion model for diverse human motion prediction},
  author={Yu, Hua and Hou, Yaqing and Pei, Wenbin and Ong, Yew-Soon and Zhang, Qiang},
  journal={IEEE Transactions on Multimedia},
  volume={27},
  pages={1848--1859},
  year={2024},
  publisher={IEEE}
}

@inproceedings{pinyoanuntapong2024controlmm,
  title={Maskcontrol: Spatio-temporal control for masked motion synthesis},
  author={Pinyoanuntapong, Ekkasit and Saleem, Muhammad and Karunratanakul, Korrawe and Wang, Pu and Xue, Hongfei and Chen, Chen and Guo, Chuan and Cao, Junli and Ren, Jian and Tulyakov, Sergey},
  booktitle={Proceedings of the IEEE/CVF International Conference on Computer Vision},
  pages={9955--9965},
  year={2025}
}

@article{huang2025self,
  title={Self forcing: Bridging the train-test gap in autoregressive video diffusion},
  author={Huang, Xun and Li, Zhengqi and He, Guande and Zhou, Mingyuan and Shechtman, Eli},
  journal={Proceedings of the Advances in Neural Information Processing Systems},
  volume={38},
  pages={167283--167308},
  year={2026}
}

@article{ma2025unitok,
  title={Unitok: A unified tokenizer for visual generation and understanding},
  author={Ma, Chuofan and Jiang, Yi and Wu, Junfeng and Yang, Jihan and Yu, Xin and Yuan, Zehuan and Peng, Bingyue and Qi, Xiaojuan},
  journal={Proceedings of the Advances in Neural Information Processing Systems},
  volume={38},
  pages={129274--129297},
  year={2026}
}

@ARTICLE{gao2025jointly,
  author={Gao, Xuehao and Yang, Yang and Du, Shaoyi and Qi, Guo-Jun and Han, Junwei},
  journal={IEEE Transactions on Multimedia}, 
  title={Jointly Understand Your Command and Intention: Reciprocal Co-Evolution Between Scene-Aware 3D Human Motion Synthesis and Analysis}, 
  year={2025},
  volume={27},
  number={},
  pages={8900-8913},}

@ARTICLE{tang2025video,
  author={Tang, Yunlong and Bi, Jing and Xu, Siting and Song, Luchuan and Liang, Susan and Wang, Teng and Zhang, Daoan and An, Jie and Lin, Jingyang and Zhu, Rongyi and Vosoughi, Ali and Huang, Chao and Zhang, Zeliang and Liu, Pinxin and Feng, Mingqian and Zheng, Feng and Zhang, Jianguo and Luo, Ping and Luo, Jiebo and Xu, Chenliang},
  journal={IEEE Transactions on Circuits and Systems for Video Technology}, 
  title={Video Understanding With Large Language Models: A Survey}, 
  year={2026},
  volume={36},
  number={2},
  pages={1355-1376},
}

@article{van2008visualizing,
  author  = {Laurens van der Maaten and Geoffrey Hinton},
  title   = {Visualizing Data using t-SNE},
  journal = {Journal of Machine Learning Research},
  year    = {2008},
  volume  = {9},
  number  = {86},
  pages   = {2579--2605},
  url     = {http://jmlr.org/papers/v9/vandermaaten08a.html}
}

\vfill

\end{document}